\documentclass[letterpaper, 10 pt, conference]{ieeeconf}  % Comment this line out if you need a4paper

\usepackage{amsmath, amssymb, amsfonts}
\usepackage[mathscr]{eucal}
\usepackage{gensymb}

\usepackage{graphicx}
\graphicspath{{./figures/}}
\usepackage[export]{adjustbox}
\usepackage{rotating}
\usepackage{float}
\usepackage{subcaption}
\usepackage{pgfplots}

\usepackage{booktabs}
\usepackage{makecell}
\usepackage{multirow}
\usepackage{tabularx}
\usepackage{hhline}
\usepackage{colortbl}
\usepackage{arydshln}

\usepackage{algorithm}
\usepackage{algpseudocode}

\let\labelindent\relax
\usepackage{enumitem}
\usepackage{multicol}
\usepackage[normalem]{ulem}
\usepackage{caption}
\usepackage{comment}

\usepackage{xcolor}

\usepackage{hyperref}
\usepackage{url}

\usepackage[colorinlistoftodos]{todonotes}

\colorlet{custom_yellow}{green!10!orange}

\definecolor{purple}{rgb}{0.56,0.27,0.68}
\definecolor{red}{rgb}{0.95,0.4,0.4}
\definecolor{purered}{rgb}{1,0,0}
\definecolor{blue}{rgb}{0.4,0.4,0.95}
\definecolor{darkblue}{rgb}{0,0,0.8}
\definecolor{grey}{rgb}{0.6,0.6,0.6}

\definecolor{col1}{RGB}{232,161,148}
\definecolor{col11}{RGB}{255,228,228}
\definecolor{col2}{RGB}{148,187,232}
\definecolor{col33}{RGB}{206,239,255}
\definecolor{col3}{RGB}{233,255,245}

\definecolor{lightgrey}{rgb}{0.85,0.85,0.85}
\definecolor{lightlightgrey}{rgb}{0.9,0.9,0.9}
\definecolor{verylightBG}{rgb}{0.9,0.99,0.99}

\definecolor{darkgreen}{rgb}{0,0.85,0.5}

\IEEEoverridecommandlockouts                              % This command is only needed if 
\title{\LARGE \bf
Bridging Handheld and Teleoperated Supervision for Contact-Rich Manipulation via State-Gated Experts
}

\author{Vidullan Surendran$^1$, Neehar Peri$^{1,2}$, David Watkins$^1$ \\ 
RAI Institute$^1$, Carnegie Mellon University$^2$}

\begin{document}

\maketitle
\thispagestyle{empty}
\pagestyle{empty}

\newcommand{\method}{Bi-modal Routing for Imitation Data via Gated Experts}
\newcommand{\acronym}{BRIDGE}
%%%%%%%%%%%%%%%%%%%%%%%%%%%%%%%%%%%%%%%%%%%%%%%%%%%%%%%%%%%%%%%%%%%%%%%%%%%%%%%%
\begin{abstract}
Handheld data collection systems, such as the Universal Manipulation Interface (UMI), enable scalable data collection across diverse environments but only capture observed actions rather than the desired actions executed by a robot controller. In contrast, teleoperation captures desired actions directly, but requires a robot in-the-loop. We revisit this trade-off through the lens of \textit{action validity} across task phases. We observe that handheld trajectories provide valid supervision in tolerant, free-space phases, but can be poor supervision targets for contact-sensitive phases, requiring high stiffness to minimize tracking errors, resulting in large contact forces. We study the interaction between these two supervision types for contact-rich manipulation and find that training policies that combine handheld data with a small number of targeted teleoperated demonstrations provide an effective hybrid strategy. Specifically, rather than teleoperating the entire task, we only collect partial teleoperated demonstrations for task segments where base handheld policies fail. However, naively mixing handheld and teleoperated phase-specific data yields worse performance than training on handheld data alone. To address this mismatch between observed and desired supervision, we propose \method \ (\acronym), a mixture of diffusion policy experts that routes between specialist task phase heads conditioned on the current robot state. Notably, our approach enables task-phase specific use of desired actions during contact sensitive segments and improves success rates over handheld-only baselines by up to 36.7\% across three contact-rich manipulation tasks. Please see our \href{https://nperi-rai.github.io/bridge-project/}{project page} for accompanying videos.
\end{abstract}

\section{Introduction}
\label{sec:intro}
The Universal Manipulation Interface (UMI)~\cite{chi2024universal} enables scalable, in-the-wild data collection by decoupling the collection process from specific robot hardware. However, this portability introduces an embodiment gap during on-robot policy execution, particularly for tasks involving contact forces, friction, and compliance. In contrast, teleoperation systems such as GELLO~\cite{wu2023gello} and ALOHA~\cite{zhao2023learning} provide high-fidelity, on-robot data but sacrifice collection speed and dataset diversity, and often lack proprioceptive feedback. 

\begin{figure}[t]
    \centering
    \includegraphics[width=\linewidth]{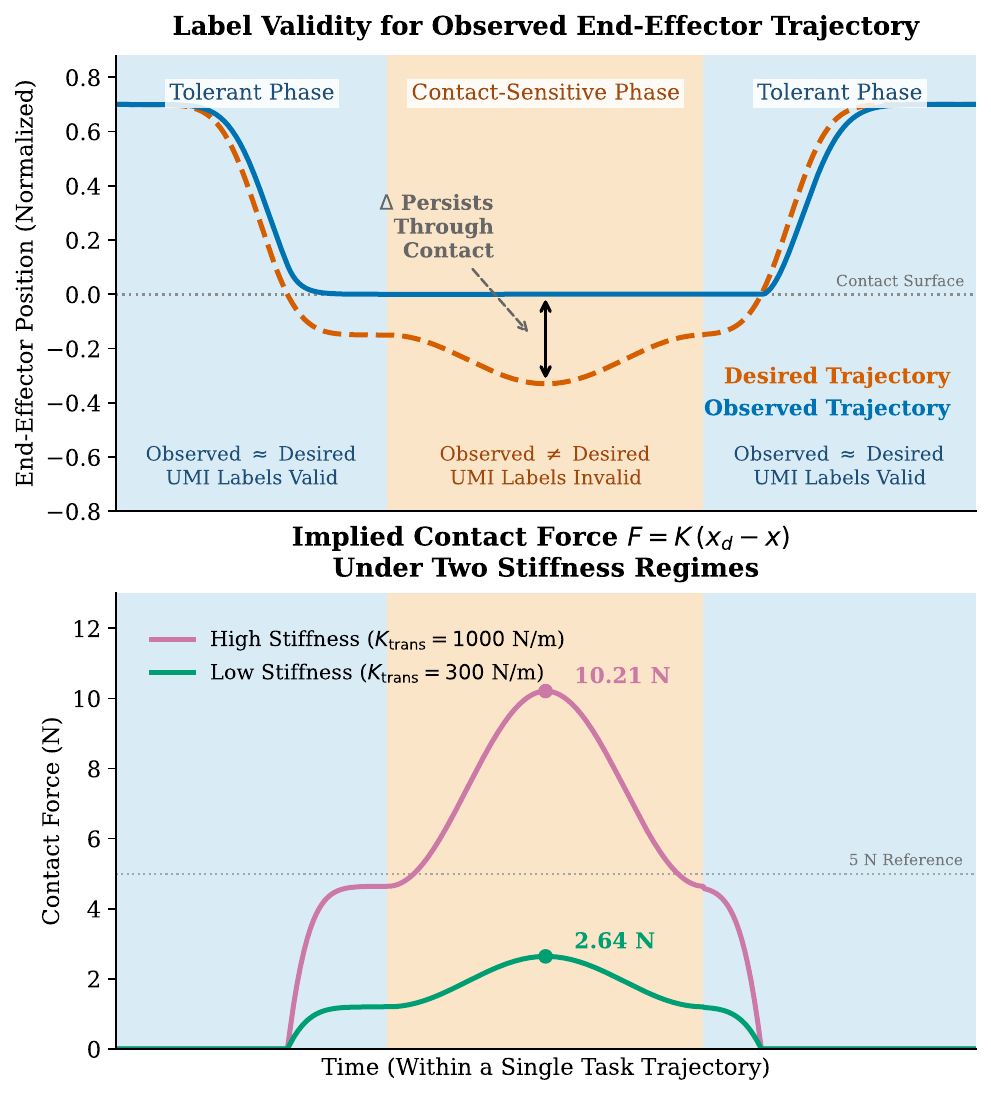}
    \caption{\small\textbf{Action Validity Under Contact.} We illustrate the end-effector trajectory and contact forces during the NIST pulley routing task~\cite{nist_assembly_task_boards}.
    \emph{Top:} In tolerant phases (blue), the observed action closely tracks the desired action. In contact-sensitive phases (yellow), the desired action drives below the contact surface; the resulting persistent error ($\Delta$) makes the observed trajectory an invalid proxy for desired intent.
    \emph{Bottom:} Proportional control converts $\Delta$ into contact force via $F \propto  K(x_d - x)$. High stiffness gains reduce tracking error but result in large contact forces, whereas lowering stiffness reduces peak force at the cost of increased tracking error.
    }
    \label{fig:cmd_obs_delta}
    \vspace{-6mm}
\end{figure}

In this paper, we examine the trade-off between handheld and teleoperated data through the lens of \textit{action validity} across different task phases. To do so, we highlight the distinction between \textit{observed} and \textit{desired} actions. An observed action is the physically realized state of the robot end-effector (or handheld device), while a desired action is the reference trajectory sent to a robot controller. Due to finite stiffness, filtering, tracking errors, and contact interactions, observed actions often deviate from desired actions. Our key insight is that handheld data (which only captures observed actions) provides valid supervision in tolerant, free-space motion but can be inadequate in contact-sensitive phases. In these contact-rich regimes, naively tracking observed actions at high stiffness generates large contact forces (Figure \ref{fig:cmd_obs_delta}). Unlike prior work that typically trains only on either handheld or teleoperated data, we demonstrate that augmenting handheld data with targeted teleoperated demonstrations (which capture both observed and desired actions) improves policy success during contact-sensitive task phases.% without significantly increasing data-collection effort. - cant defend this since we don't havce detailed collection stats with controlled rest periods. Mayber rephasre to reduced compated to full teleop?

\textbf{Bridging the Embodiment Gap.} Despite their complementary strengths, naively combining handheld and teleoperated data often degrades policy performance because observed and desired actions do not always provide compatible supervision. To resolve this mismatch, we propose \method \ (\acronym), a mixture of diffusion policy experts that routes between specialized task heads conditioned on the current robot state. This framework establishes a deliberate division of labor: cheap handheld data helps learn the task scaffold, while teleoperation is reserved for contact-rich bottlenecks where observed actions fail to provide reliable supervision. We find that sparse teleoperation at contact bottlenecks approaches the performance of full-task teleoperation while reducing teleoperation data collection effort.

\textbf{Contributions.} We present three primary contributions. First, we study \textit{action validity} in data collection, demonstrating that observed actions are sufficient for error-tolerant, free-space motion but are unreliable during contact-sensitive phases, where desired action supervision substantially improves execution. Second, we introduce \method, a mixture of diffusion policy experts that jointly learns from data sources containing both observed and desired action labels. Third, we show that our approach naturally supports task-phase aware execution, allowing contact sensitive segments to use desired action supervision to improve task success rates over handheld-only baselines by up to 36.7\% across three contact-rich manipulation tasks.

\section{Related Works}

\textbf{Hybrid Data Collection Strategies.} Recent work increasingly leverages data across diverse embodiments and modalities to improve policy generalization. Large-scale efforts like Open X-Embodiment~\cite{openx2023} and DROID~\cite{khazatsky2024droid} aggregate demonstrations at scale to train foundational policies~\cite{team2024octo, intelligence2025pi_}. Other approaches address data heterogeneity by routing through embodiment-specific heads~\cite{wang2024hpt}, closing morphological gaps via rendering or human videos~\cite{fu2024humanplus}, and co-training across the sim-to-real boundary~\cite{torne2024rialto}. Alternatively, synthetic augmentation can be used to expand small demonstration sets~\cite{mandlekar2023mimicgen}. However, naively mixing heterogeneous human demonstrations can degrade performance compared to using a single source~\cite{mandlekar2021matters}. Unlike prior work that aggregates disparate data under a shared action space, we mix two distinct types of supervision from the \textit{same robot embodiment}. Specifically, our handheld data only provides observed actions, while the teleoperated data provides desired actions, allowing us to route between observed and desired trajectories based on local action validity.

\textbf{Learning from Corrective Behaviors.} Collecting targeted human demonstrations to correct residual errors in behavior cloning is an active area of research. Iterative methods like DAgger~\cite{ross2011reduction} query experts to label all states visited by the policy, while more recent variants reduce this burden by only querying experts in uncertain or risky states~\cite{hoque2021thrifty}. Similarly, intervention-based approaches keep a human-in-the-loop during data collection, reweighting demonstrations to emphasize corrections~\cite{spencer2020learning,liu2023sirius}. Unlike iterative approaches, Residual Policy Learning (RPL)~\cite{silver2018residual,johannink2019residual} composes a learned correction over a base controller. Recent work extends RPL to contact-rich manipulation under impedance control~\cite{xu2025compliant}. However, these methods typically collect corrections with a model in the loop. In contrast, although our targeted demonstrations are informed by model failures, we collect our teleoperated examples in a single offline session. %Our approach routes between action experts trained on either handheld or teleoperated supervision.

\textbf{Mixture-of-Experts Policies.} Mixture-of-Experts (MoE) architectures, which activate sparse subnetworks per input, have been instrumental in scaling language models~\cite{shazeer2017outrageously}. Recently, sparse MoE diffusion policies~\cite{hao2026abstracting} have been adapted for contact-rich manipulation. These methods use soft routing to separate latent distributions, effectively mitigating mode averaging across heterogeneous demonstrations. Although our architecture shares the high-level intuition of routing inputs to specialized experts, we neither scale our parameter count nor train the router implicitly using a standard task loss. Instead, our experts correspond to two distinct sources of supervision.

\section{Learning a State-Gated Mixture of Experts}

\label{sec:model_arch}
In this section, we present DM-UMI, our dual-mode data collection setup for contact-rich manipulation. We then introduce \acronym, our mixture-of-experts policy architecture that learns from heterogeneous supervision. Unlike residual policies, BRIDGE does not require the {\tt support} expert to learn a correction over the full {\tt base} action distribution. Instead, it treats observed and desired actions as distinct supervision types and routes to the {\tt support} expert when in the {\tt support} action manifold.

\begin{figure}[t]
    \centering
    \includegraphics[trim={0.5cm 3.5cm 0cm 0cm},clip,width=\linewidth]{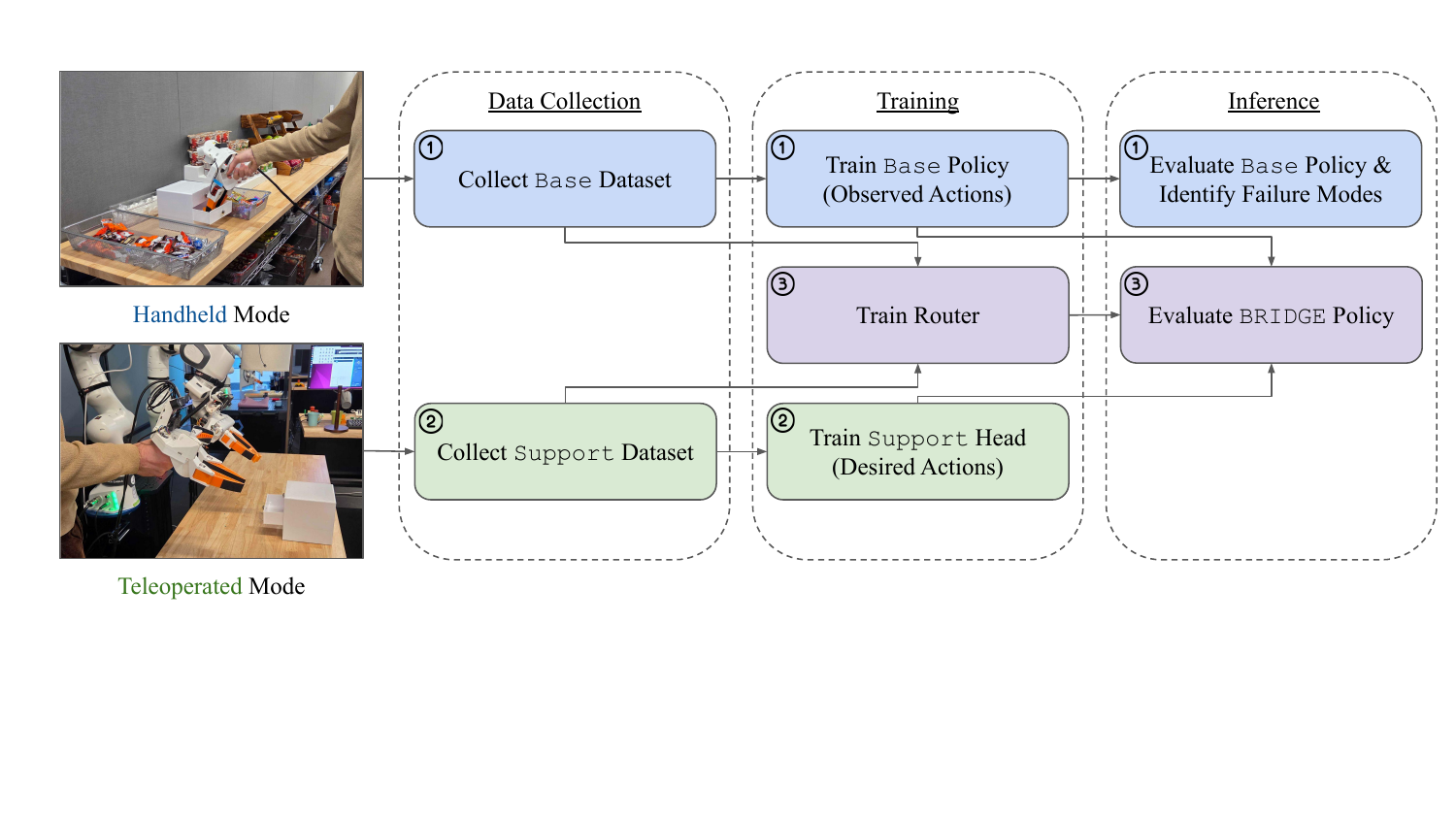}
    \caption{\small\textbf{Dual Mode Data Collection Pipeline.} First, we use DM-UMI in handheld mode to collect {\tt base} demonstrations to learn the task scaffold. We then train and evaluate this {\tt base} policy to identify failure modes. Second, we use DM-UMI in teleoperation mode to collect a targeted {\tt support} dataset to address {\tt base} policy failures. We then freeze the {\tt base} policy and train a {\tt support} head. Third, we train a router to hard-switch between the {\tt base} and {\tt support} heads at inference. Our \acronym \ model combines the {\tt base} policy, {\tt support} head, and router (Figure \ref{fig:model_arch}).
} 
    \label{fig:dual_mode}
    \vspace{-6mm}
\end{figure}

\textbf{DM-UMI: Dual-Mode Data Collection Device.} UMI \cite{chi2024universal} relies on offline monocular SLAM for trajectory reconstruction and ArUco tag detection for gripper width estimation. While this enables in-the-wild data collection, it requires offline processing to extract camera pose and gripper width. Since teleoperation requires online estimates of these parameters, we design a dual-mode handheld device called Dual-Mode UMI (DM-UMI) that modifies UMI's sensing stack to support both handheld and teleoperated data collection. DM-UMI uses an XVISIO DS80 SLAM camera to provide online 6-DoF pose estimates, and magnetic position sensors to measure gripper width. We use these pose and width values to command the robot end effector and gripper. 

\begin{figure}[t]
    \centering
    \includegraphics[trim={0cm 4cm 0cm 0cm},clip,width=\linewidth]{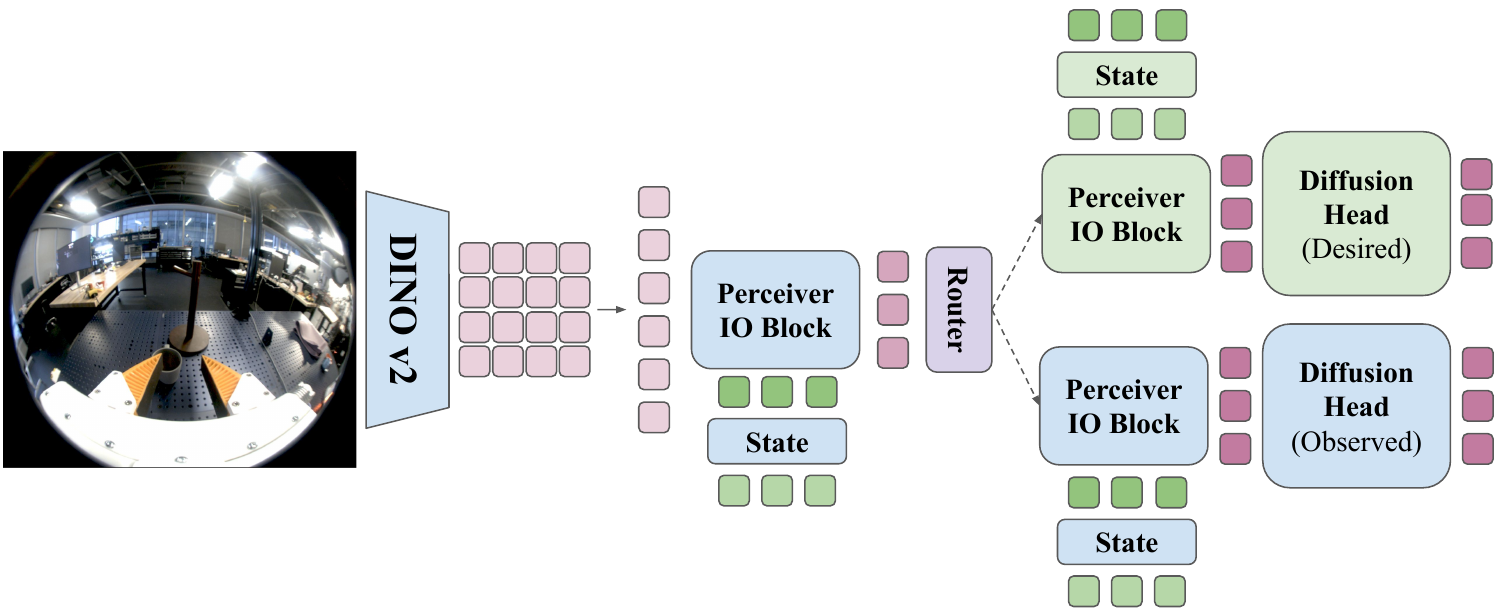}
    \caption{\small\textbf{Model Architecture.} We propose \acronym, an extension of Diffusion Policy that dynamically routes between predicting observed and desired actions. Visual observations are encoded via DINOv2 \cite{oquab2023dinov2}, processed through a Perceiver IO block, and fused with state features via cross-attention. This shared latent representation is passed to a router, which switches between the observed and desired diffusion heads. To train \acronym, we sequentially train the {\tt base} model (shown in \textcolor{blue}{blue}), collect targeted teleoperated demonstrations to train the {\tt support} head (\textcolor{darkgreen}{green}), and finally train the router (\textcolor{purple}{purple}).
}
    \label{fig:model_arch}
    \vspace{-6mm}
\end{figure}

\textbf{Dataset Collection.}
We use DM-UMI to collect both handheld and teleoperated demonstrations (Figure~\ref{fig:dual_mode}). First, we use DM-UMI to collect task demonstrations across diverse environments, referred to as the \texttt{base} dataset. We train a \texttt{Base} policy on this data and empirically identify task phases associated with systematic failures. Next, we teleoperate the robot using the DM-UMI to collect targeted demonstrations for these phases in a single pass rather than through iterative policy-in-the-loop data collection. This \texttt{support} dataset implicitly captures embodiment-specific effects, including controller dynamics, robot kinematics, contact dynamics, and grasp stability. This approach leverages the complementary structure of both datasets, where the \texttt{base} dataset (observed actions only) provides nominal task coverage, while the \texttt{support} dataset (desired actions) isolates localized regimes where the \texttt{base} policy is unreliable. We describe the \texttt{base} policy, and present our approach below (Figure \ref{fig:model_arch}).

\textbf{Stage 1: \texttt{Base} Policy Training.}
Our base policy architecture is derived from Diffusion Policy~\cite{diffusionpolicy}. Similar to prior work, we model action prediction as denoising future action chunks conditioned on visual and proprioceptive observations. However, we introduce several architectural modifications motivated by the demands of precise manipulation, multiple data streams, and heterogeneous supervision. Many diffusion-based policies \cite{diffusionpolicy, chi2024universal} extract visual features from a ViT encoder and only use the class (CLS) token as a global visual descriptor. While efficient, this compresses spatial structure early in the model, making it difficult to reason about the spatial relationship between the end effector and the environment for precise, contact-rich tasks. Therefore, we retain the full set of spatial patch tokens produced by the DINOv2 \cite{oquab2023dinov2} vision encoder. For a batch size $B$, a backbone with patch size $N$, and a latent dimension $D$, an image is encoded into $V \in \mathbb{R}^{B \times N \times D}$ tokens. Given a 518 $\times$ 518 image and a patch size of 14, DINOv2 produces 1369 spatial tokens. Therefore, we employ a PerceiverIO-style \cite{jaegle2021perceiver} aggregation to reduce the number of tokens with a set of learnable queries,  $Q\in \mathbb{R}^{B\times M \times D}, M \ll N$ via stacked cross-attention blocks $Z_{vision}=XATTN(Q,V)$. This information bottleneck effectively extracts task-relevant structure while discarding irrelevant visual details. Similarly, state inputs such as end-effector pose and gripper width are concatenated and linearly projected into $\mathbb{R}^{D}$ as $Z_{state}$ and are cross-attended with the vision tokens to generate the latent conditioning $Z_{latent} = XATTN(Z_{state}, Z_{vision})$. We refer to the modules that generate $Z_{vision}$, $Z_{state}$ and $Z_{latent}$ as the latent adapter $\phi_b$. We follow the standard conditional denoising formulation used in Chi et al. ~\cite{diffusionpolicy} where $Z_{latent}$ is used as input for a temporal diffusion policy, and train with standard diffusion losses on the \texttt{base} dataset. 

\textbf{Stage 2: \texttt{Support} Head Training.}
With the {\tt base} expert $\pi_b$ and shared vision encoder kept frozen, we train the {\tt support} latent adapter  $\phi_s$, and head $\pi_s$ on the {\tt support} dataset. The action target for this expert is the desired trajectory, allowing it to map the latent to a desired action chunk, $\hat{a}^{s}_{t:t+H} = \pi_s(\phi_s(Z_{latent}))$.  The support expert is optimized independently with the standard diffusion loss.

\textbf{Stage 3: Router Training.} Finally, we train a MLP gate $G_\psi(z)$ to dynamically route observations to the appropriate expert. To train this gate, we extract the intermediate $Z_{latent}$ from $\pi_b$ for samples in both the \texttt{base} and \texttt{support} datasets. To avoid ambiguous pseudo-labels in regions where base and support demonstrations overlap, we remove {\tt base} latents within an $\epsilon$-neighborhood of \texttt{support} latents before constructing the router training set. Given the filtered {\tt base} latents $B_b$ and \texttt{support} latent bank $B_s$, we compute the similarity to the $k=16$ neighbors in each bank,
\begin{equation}
\begin{split}
    \sigma_{+}(z) = \frac{1}{k}\sum_{z_s \in \mathcal{N}_k(z, B_s)} \cos(z,z_s), \\
    \sigma_{-}(z) = \frac{1}{k}\sum_{z_b \in \mathcal{N}_k(z, B_b)} \cos(z,z_b)
\end{split}
\end{equation}
To account for overlap between the local neighborhoods, we use a class relative score with a decision threshold $\eta_{\rho}$ to generate the label $\rho(z)$,
\begin{equation}
    \rho(z) = (\sigma_{+}(z) - \sigma_{-}(z)) > \eta_{\rho},
\end{equation}
Lastly, we distill this $k$-NN classifier into $G_\psi$ using a binary cross-entropy loss, and train it to predict $\rho(z)$.

\textbf{Router Architecture.} The router follows the same structure as the \texttt{base} policy with the action head replaced by a lightweight residual MLP. We reuse the DINOv2 tokens and the state input from the \texttt{base} policy. The state inputs are linearly projected to the latent dimension of $d=384$, and fused with the vision tokens using a PerceiverIO-style aggregation. The output tokens are layer normalized and passed through a two-layer MLP with GELU activation. The final linear layer of the adapter is zero-initialized, so that the adapter initially acts as an identity mapping.

\textbf{\acronym \ Inference.}
During inference, each observation $o$ at time $t$ is encoded into shared visual and state features. The {\tt base} and {\tt support} adapters produce expert-specific latents $z^b_t=\phi_b(Z_{vision},Z_{state})$ and $z^s_t=\phi_s(Z_{vision},Z_{state})$. The router predicts support probabilities $g_{t:t+H}=G_\psi(z^b_t)$. Given an inference threshold $\eta$, we hard switch via,
\begin{equation}
    \hat{a}^{b}_{t+j} = \pi_b(z^b_t), \qquad \hat{a}^{s}_{t+j} = \pi_s(z^s_t)
\end{equation}
\begin{equation}
    m_{t+j} = \mathbb{I}[g_{t+j} > \eta], \qquad \hat{a}_{t+j} = (1-m_{t+j})\hat{a}^{b}_{t+j} + m_{t+j} \hat{a}^{s}_{t+j}
\end{equation}

\textbf{Why Use Hard-Switching for Routing?}
During inference, the router $G_\psi(z_t)$ selects between the {\tt base} and {\tt support} experts using a hard switch. We use hard switching rather than action blending because the two experts may represent distinct behavioral modes. The {\tt base} policy is trained on observed actions, while the {\tt support} expert is trained on desired actions, as discussed in Section~\ref{sec:intro}. Averaging actions from these experts can therefore produce actions that are not meaningful for either mode. Although hard switching can introduce boundary sensitivity, the downstream control stack smooths switching discontinuities. The selected policy output is passed to a trajectory generator that interpolates commanded motion at 100~Hz, which is then streamed to a 1~kHz Cartesian impedance controller with torque smoothing. During rollouts, we damp transient switching by adding temporal hysteresis that activates the support expert only when $G_\psi(z_t) > \eta$, with $\eta=0.5$, for a continuous duration of 0.25~seconds.

\section{Experiments}
We evaluate baseline policies on three contact-rich manipulation tasks and demonstrate that \acronym~consistently outperforms handheld baselines and approaches the performance of policies trained on full teleoperated data. Lastly, we characterize the performance of our router.

\textbf{Dual-Mode System Characterization.} 
We evaluate the DS80's pose tracking accuracy in Table \ref{tab:tracking_accuracy}. To characterize the behavior of our dual-mode data collection system, we collected 1000 DM-UMI demonstrations in a motion capture environment (20 Vicon Vantage V16 cameras). The system exhibits a mean absolute translation error of 25.9 mm and a mean absolute rotation error of 6.3 degrees. However, its precision improves substantially when measuring relative poses. The mean relative translation error decreases to 1.1 mm, while the relative rotation error drops to 0.2 degrees, which is a more relevant measure for our policies, since they are trained to predict relative poses. %We plan to release this dataset to support future research on handheld SLAM evaluation.  

{
\setlength{\tabcolsep}{0.75em}
\begin{table}[b]
\centering
\caption{\textbf{DS80 Pose Tracking Accuracy.} We collect a dataset of 1000 DM-UMI demonstrations interacting with diverse objects in a motion capture room to precisely evaluate the DS80's absolute and relative 6-DoF pose error. Across all demonstrations, we find that the DS80 has an average absolute translation error of 25.9 mm and an absolute rotation error of 6.3 degrees. However, since we train our policies with relative poses, tracking error drops significantly.}
\label{tab:tracking_accuracy}
\small
\resizebox{0.95\linewidth}{!}{
\begin{tabular}{lcccc}
\toprule
\textbf{} & \textbf{Abs. Translation} & \textbf{Rel. Translation} & \textbf{Abs. Rotation} & \textbf{Rel. Rotation} \\
 & \textbf{(mm)} & \textbf{(mm)} & \textbf{(deg)} & \textbf{(deg)} \\
\midrule
Median & 22.6 & 0.9 & 4.0 & 0.2  \\
Mean   & 25.9 & 1.1 & 6.3 & 0.2  \\
STD    & 16.1 & 0.7 & 5.1 & 0.2  \\
\bottomrule
\end{tabular}}
% \vspace{-6mm}
\end{table}}

{
\setlength{\tabcolsep}{0.75em}
\begin{table}[t]
\centering
\caption{\textbf{Comparison of Episode Duration.} We present a comparison of the average time of a single demonstration (seconds $\pm$ std-dev) with our DM-UMI device for handheld, targeted teleoperation, and full-task teleoperation respectively. Full task teleoperated demonstrations were collected under a comparable data collection time budget, which included task resets and collector breaks.} 
%However, the due to the difficulty of teleoperating long-horizon tasks, teleoperation took approximately 3 $\times$ longer wall clock time than our hybrid approach.
\label{tab:collection_time}
\small
\resizebox{\linewidth}{!}{
\begin{tabular}{lcccc}
\toprule
\textbf{Task} & \textbf{Handheld} & \textbf{Targeted Teleoperation} & \textbf{Full-Task Teleoperation} \\
\midrule

Pipe Insertion & $6.7 \pm 1.6$ & $6.7 \pm 2.0$ & $15.3 \pm 2.2$ \\
Battery Insertion & $11.1 \pm 1.7$ & $15.5 \pm 3.6$ & $27.6 \pm 4.0$ \\
NIST Pulley Task & $10.5 \pm 2.6$ & $9.7 \pm 4.5$ & $24.7 \pm 3.3$ \\

\bottomrule
\end{tabular}}
\vspace{-3mm}
\end{table}}

We find that handheld DM-UMI demonstrations are faster to collect than teleoperated ones (Table ~\ref{tab:collection_time}). Across all evaluated tasks, handheld collection is roughly 3$\times$ faster than full-task teleoperation, and partial teleoperation is roughly 2$\times$ faster. However, these runtime statistics still underestimate the practical cost of full teleoperation, which is more physically tiring. This cost is also highly task-dependent. For short, spatially compact tasks like pipe insertion, experienced operators collect full-task demonstrations efficiently. However, this efficiency degrades in longer-horizon tasks requiring large workspace motions and sustained contact. In pulley routing, for example, the operator must maintain O-ring tension over a wide range of motion without direct force feedback. We found this substantially more difficult; not only was the operator forced into awkward postures, but the robot safety limit faults were triggered frequently during collection. In contrast, \acronym~decouples broad nominal coverage from localized teleoperation. Handheld demonstrations provide the foundational task scaffold, while teleoperated {\tt support} is collected exclusively around failure regions where observed action labels are inadequate. Therefore, even when full teleoperation provides a strong in-domain upper bound, targeted {\tt support} collection minimizes the physical burden of continuous teleoperation and eliminates the redundancy of repeatedly teleoperating trivial free-space phases.

\begin{table*}[t]
    \centering
    \caption{
    \small\textbf{Dataset Statistics.} We summarize the number of collected \texttt{base}, \texttt{support}, and fully teleoperated demonstrations below. {\tt Support} Temporal \% is the median fraction of time and {\tt support} distance \% is the median fraction of the end-effector path length covered by a support demonstration.
    }
    \label{tab:support_fraction}
    \begin{tabular*}{\textwidth}{@{\extracolsep{\fill}}lccccc}
        \toprule
        Task & {\tt Base} \# & {\tt Support} \# & {\tt Teleop} \# & {\tt Support} Temporal \% & {\tt Support}  Dist. \%\\
        \midrule
        NIST Pulley Routing & 201 & 50 & 60 & 37.8\% & 7.4\% \\
        Pipe Insertion & 100 & 81 & 60 & 24.5\% & 3.4\% \\
        Battery Insertion & 290 & 59 & 100 & 61.1\% & 14.9\% \\
        \bottomrule
    \end{tabular*}
    \vspace{-6mm}
\end{table*}

\textbf{Dataset.} We summarize the number of demonstrations collected per task in Table~\ref{tab:support_fraction}. While {\tt support} data often occupies a large fraction of the trajectories' temporal extent due to slower collection rates, it remains spatially sparse. We collect handheld demonstrations for each task until the {\tt base} policy converges during the free-space task phases. Rather than relying on exhaustive data collection, we demonstrate that targeted, small-scale {\tt support} demonstrations are sufficient to significantly improve task success. Collecting the full-task teleoperated dataset was considerably more cognitively and physically demanding than collecting the {\tt base}-plus-{\tt support} dataset.

\begin{figure*}[t]
    \centering
    \setlength{\tabcolsep}{1pt}
    \renewcommand{\arraystretch}{0}
    \begin{tabular}{@{}cccccc@{}}
        \begin{subfigure}{0.155\linewidth}\centering\includegraphics[width=\linewidth]{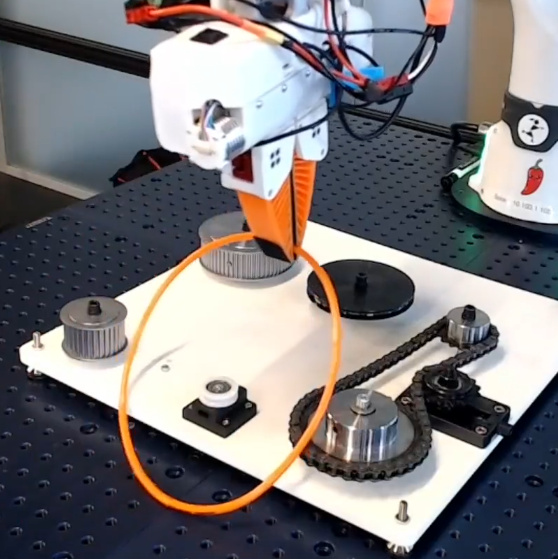}\end{subfigure} &
        \begin{subfigure}{0.155\linewidth}\centering\includegraphics[width=\linewidth]{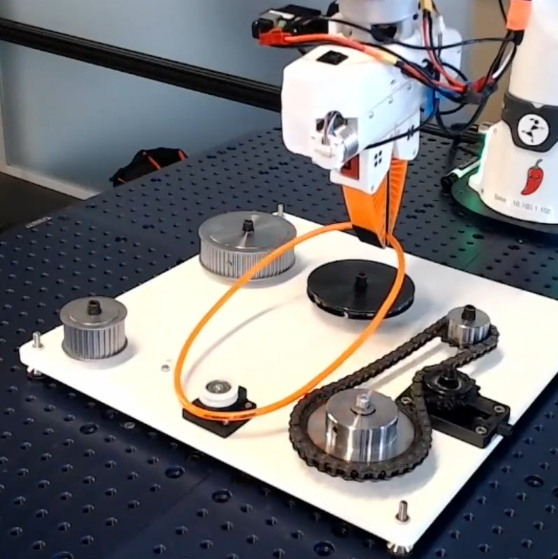}\end{subfigure} &
        \begin{subfigure}{0.155\linewidth}\centering\includegraphics[width=\linewidth]{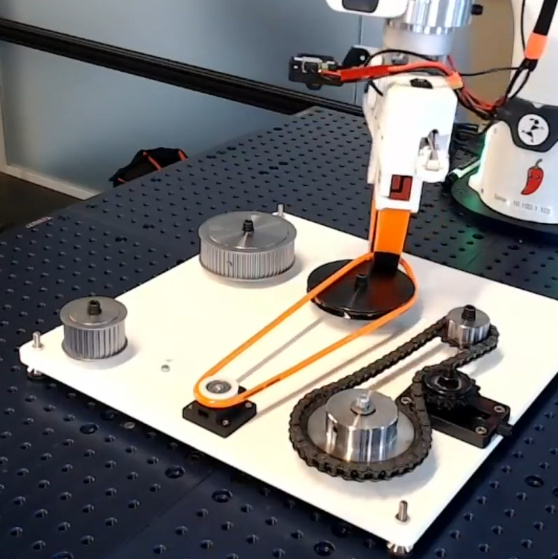}\end{subfigure} &
        \begin{subfigure}{0.155\linewidth}\centering\includegraphics[width=\linewidth]{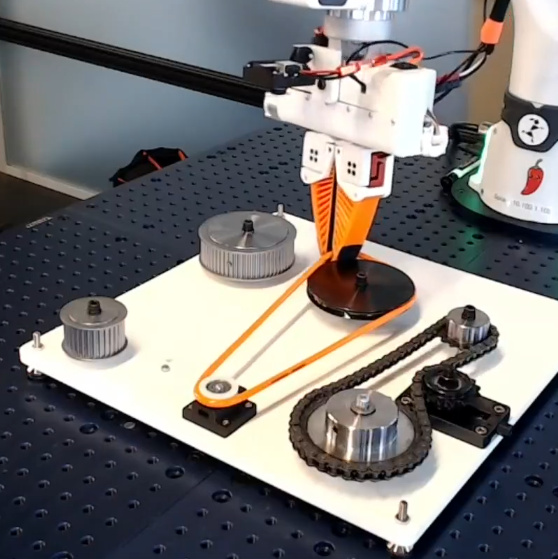}\end{subfigure} &
        \begin{subfigure}{0.155\linewidth}\centering\includegraphics[width=\linewidth]{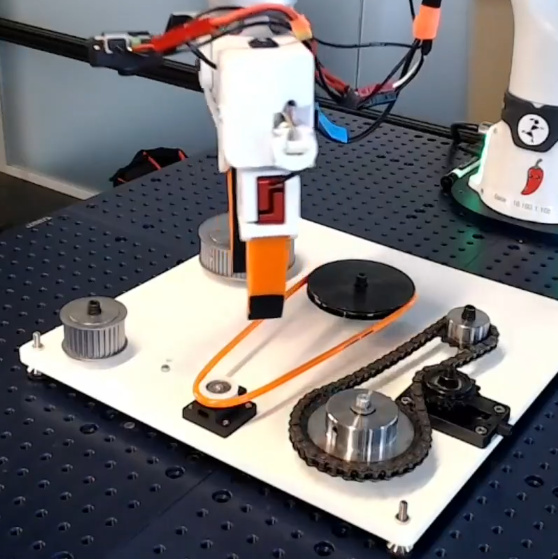}\end{subfigure} &
        \begin{subfigure}{0.155\linewidth}\centering\includegraphics[width=\linewidth]{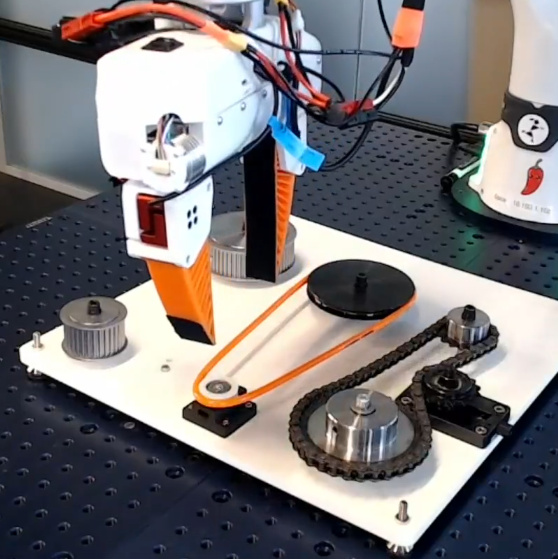}\end{subfigure} \\[2pt]
        \begin{subfigure}{0.155\linewidth}\centering\includegraphics[width=\linewidth]{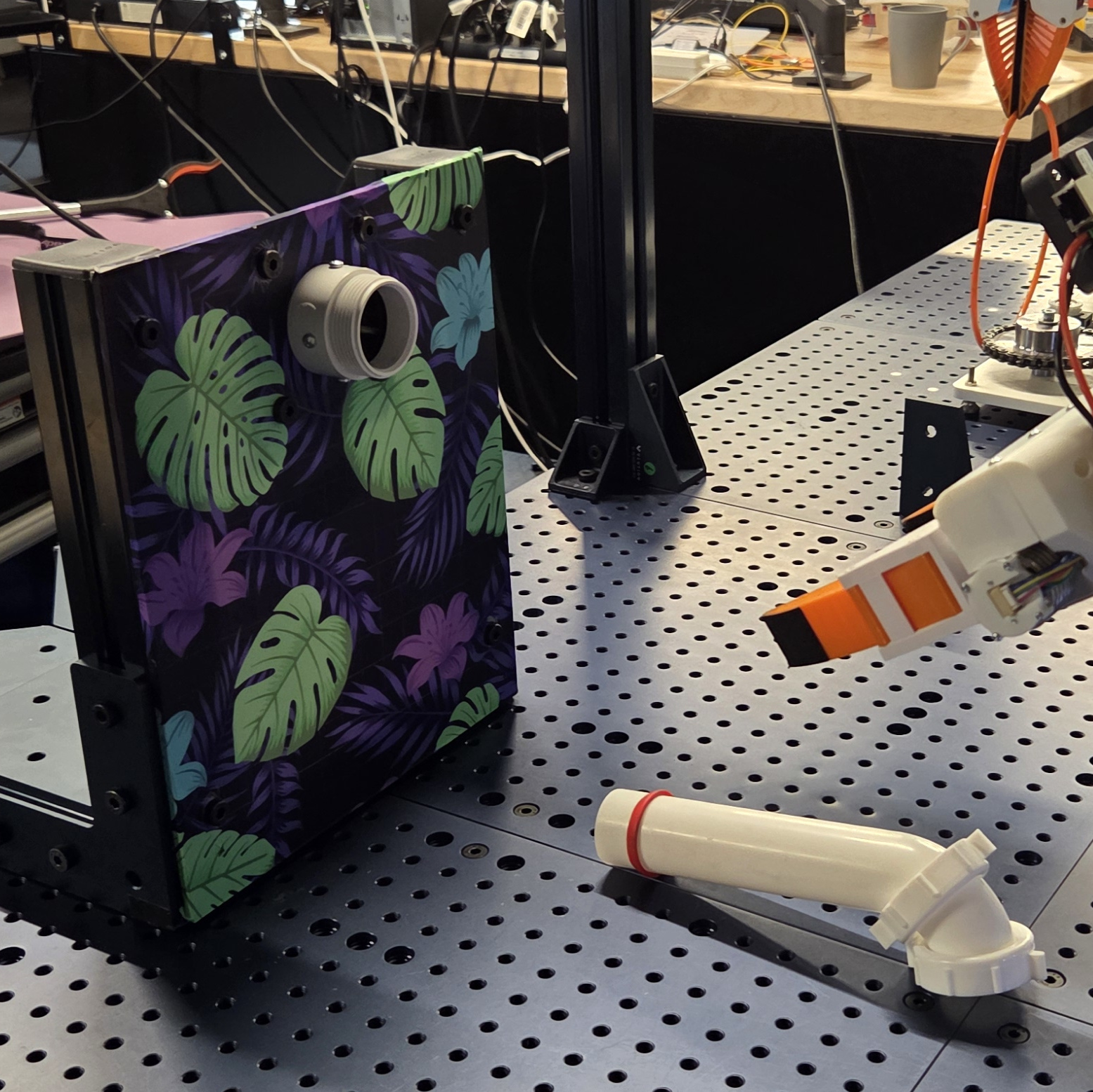}\end{subfigure} &
        \begin{subfigure}{0.155\linewidth}\centering\includegraphics[width=\linewidth]{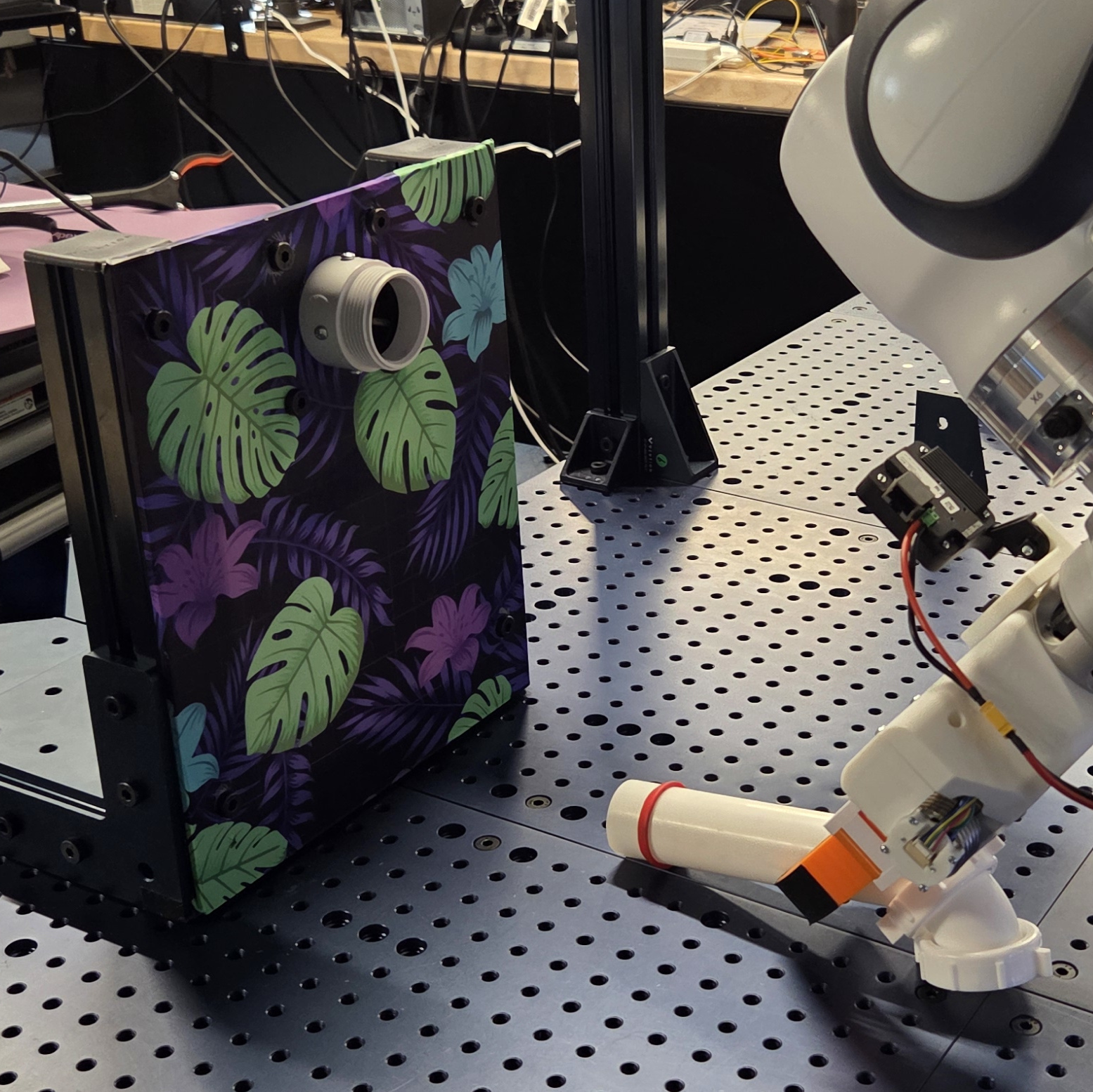}\end{subfigure} &
        \begin{subfigure}{0.155\linewidth}\centering\includegraphics[width=\linewidth]{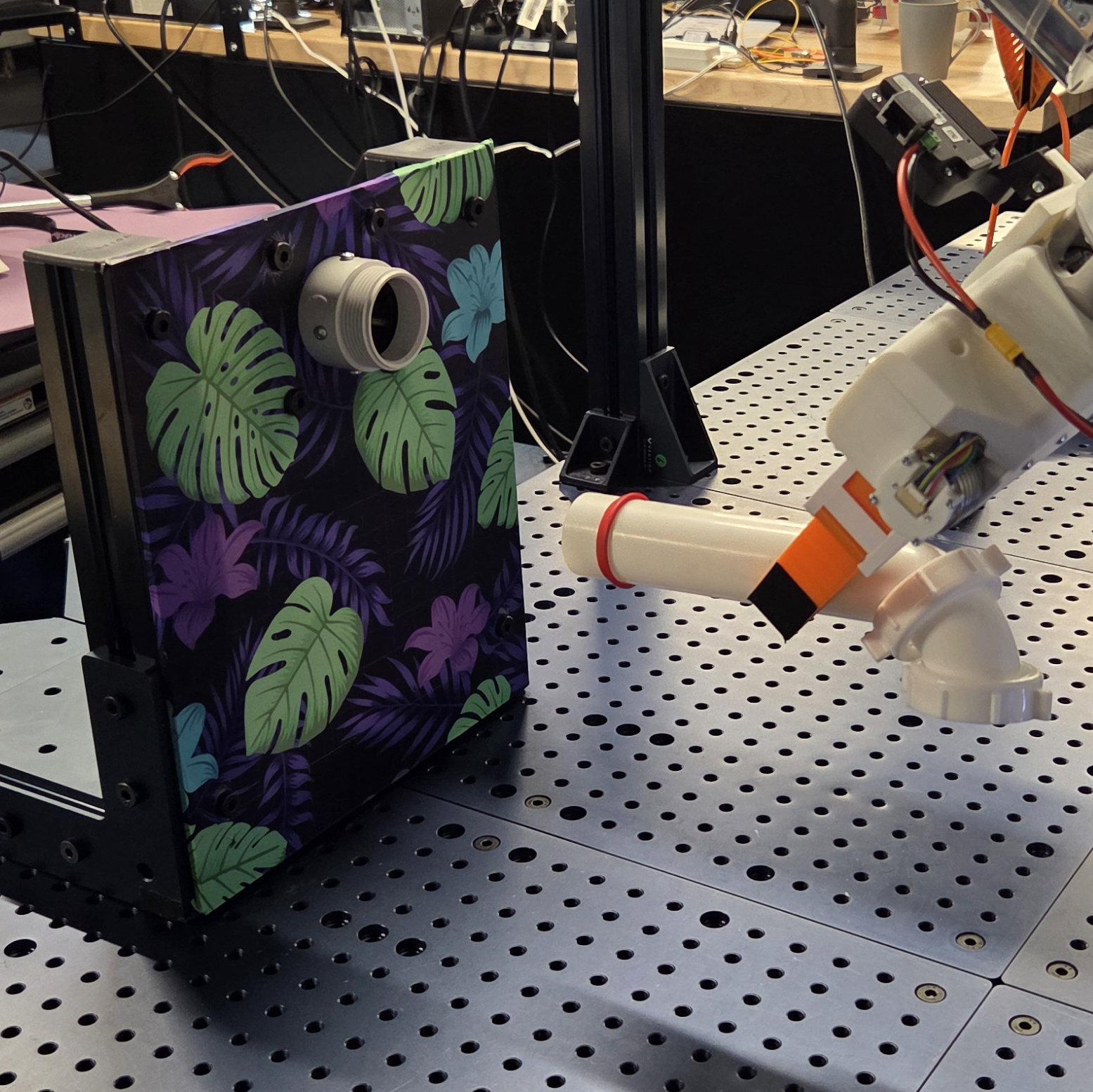}\end{subfigure} &
        \begin{subfigure}{0.155\linewidth}\centering\includegraphics[width=\linewidth]{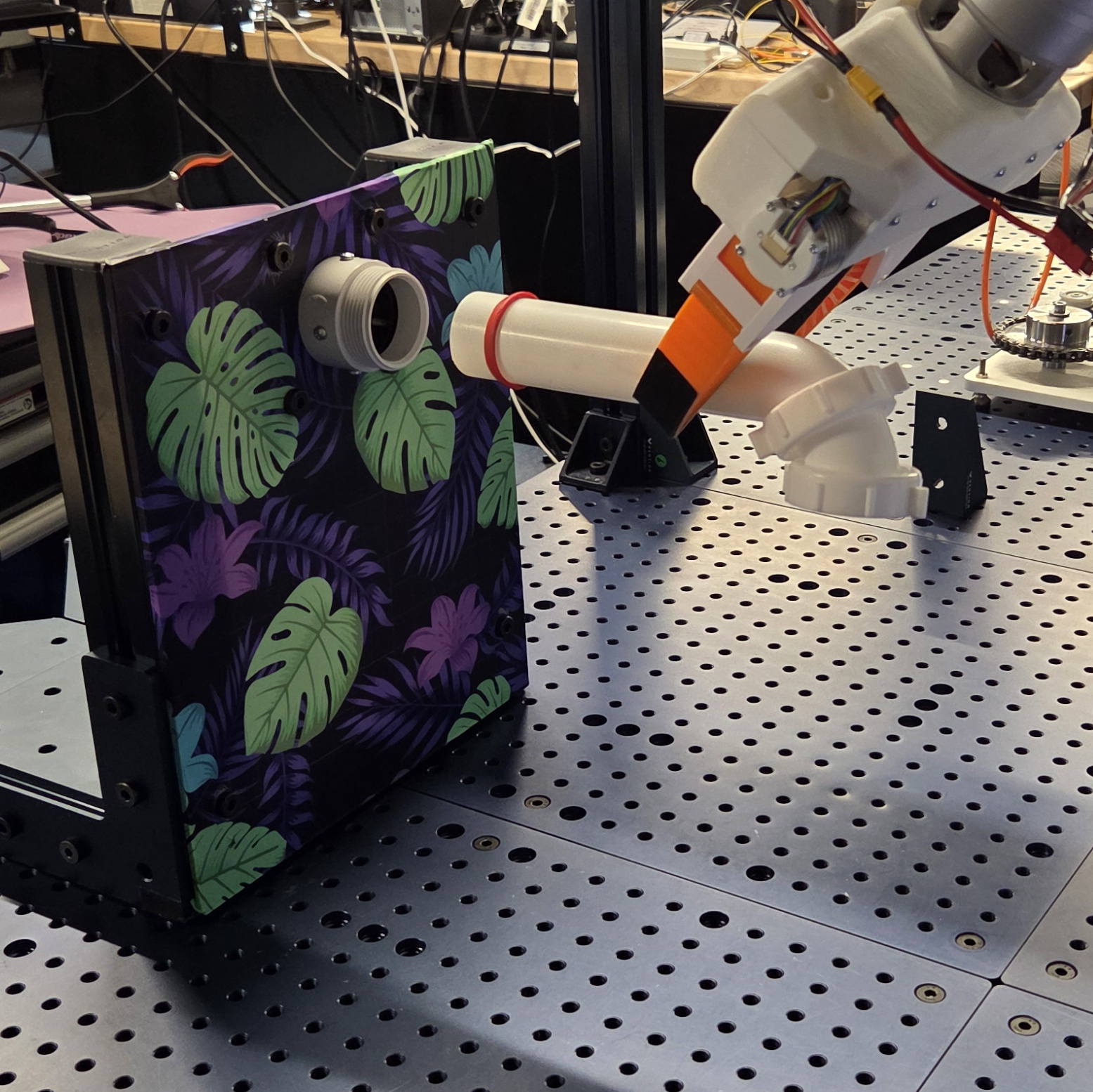}\end{subfigure} &
        \begin{subfigure}{0.155\linewidth}\centering\includegraphics[width=\linewidth]{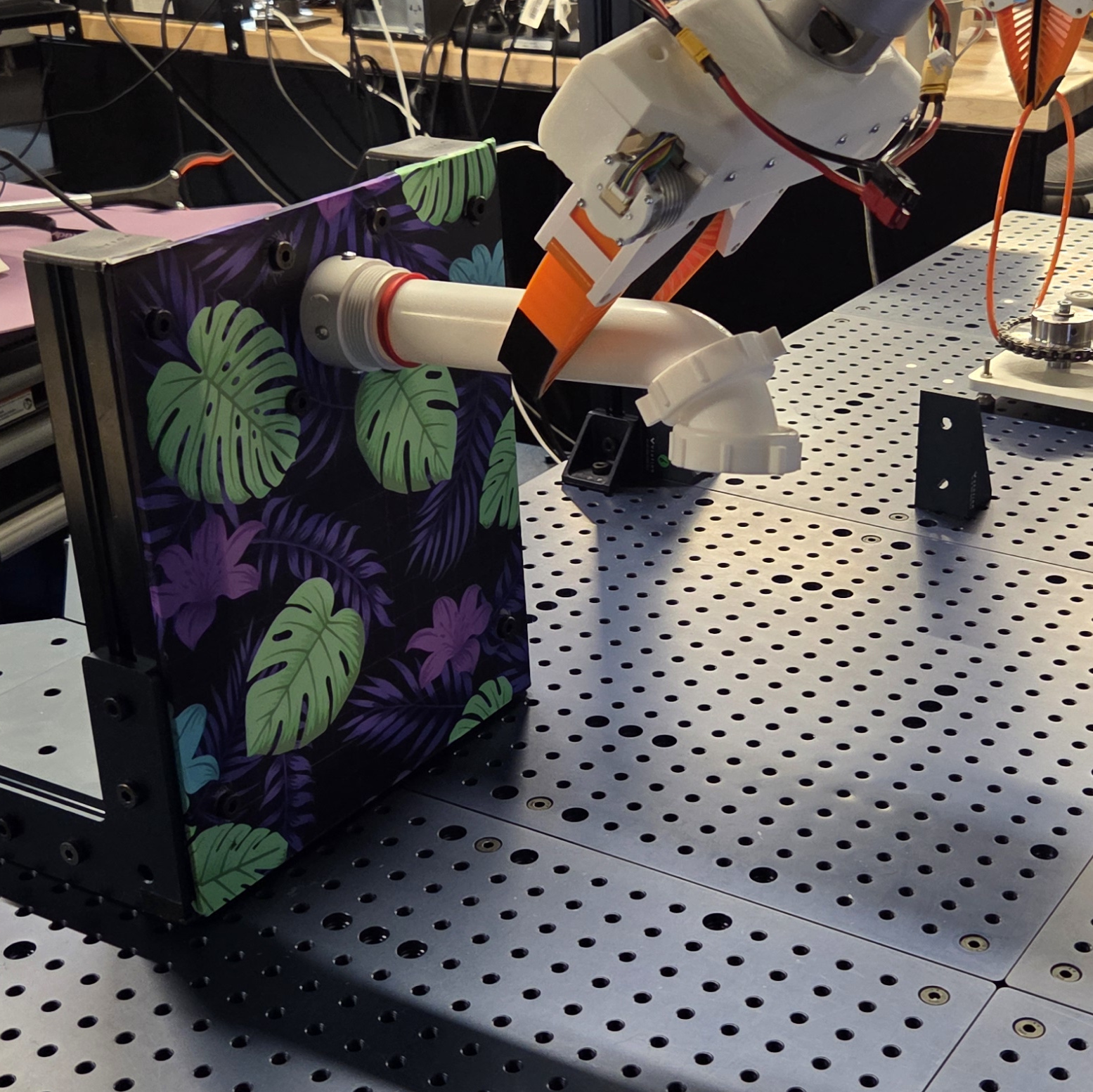}\end{subfigure} &
        \begin{subfigure}{0.155\linewidth}\centering\includegraphics[width=\linewidth]{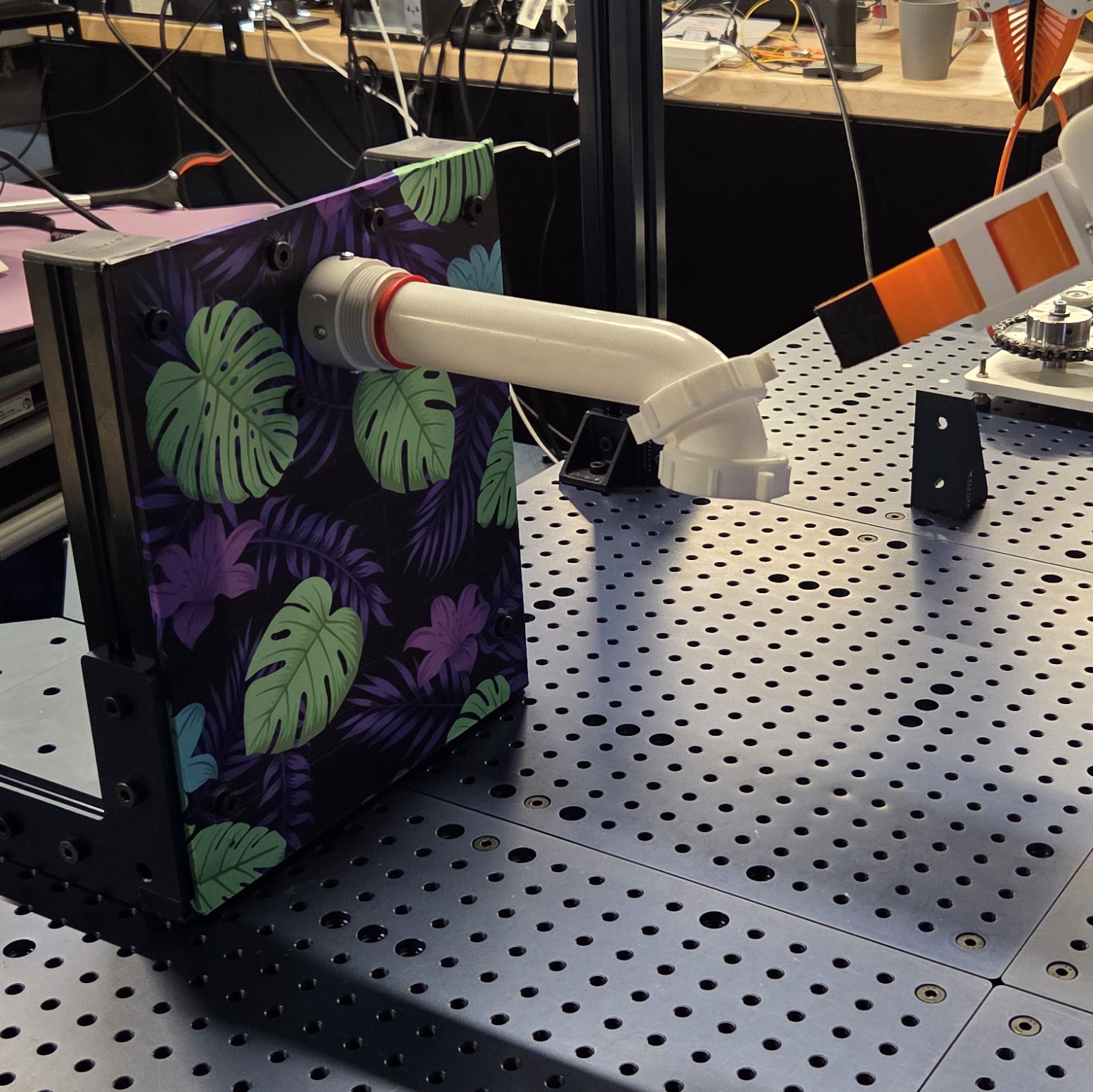}\end{subfigure} \\[2pt]
        \begin{subfigure}{0.155\linewidth}\centering\includegraphics[width=\linewidth]{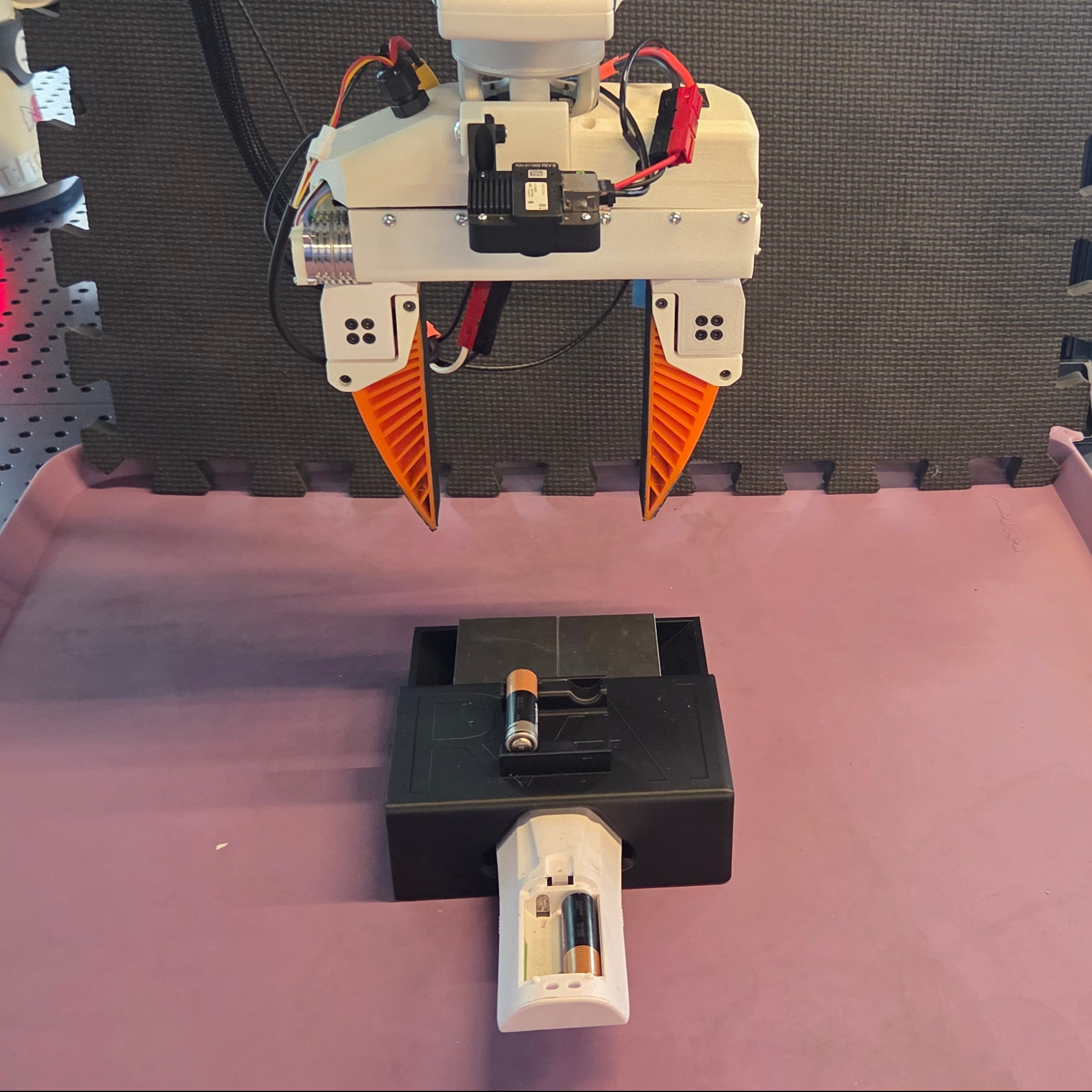}\end{subfigure} &
        \begin{subfigure}{0.155\linewidth}\centering\includegraphics[width=\linewidth]{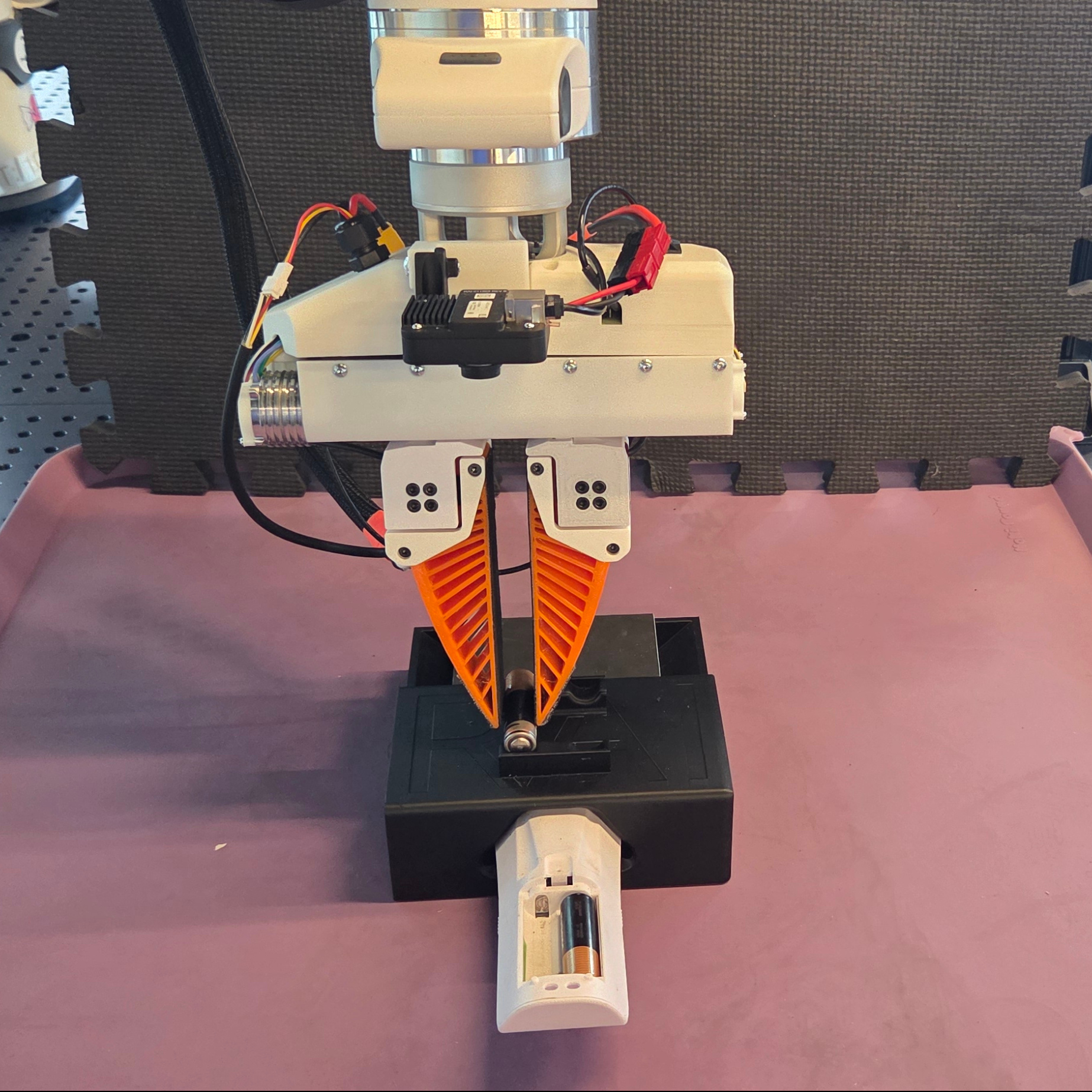}\end{subfigure} &
        \begin{subfigure}{0.155\linewidth}\centering\includegraphics[width=\linewidth]{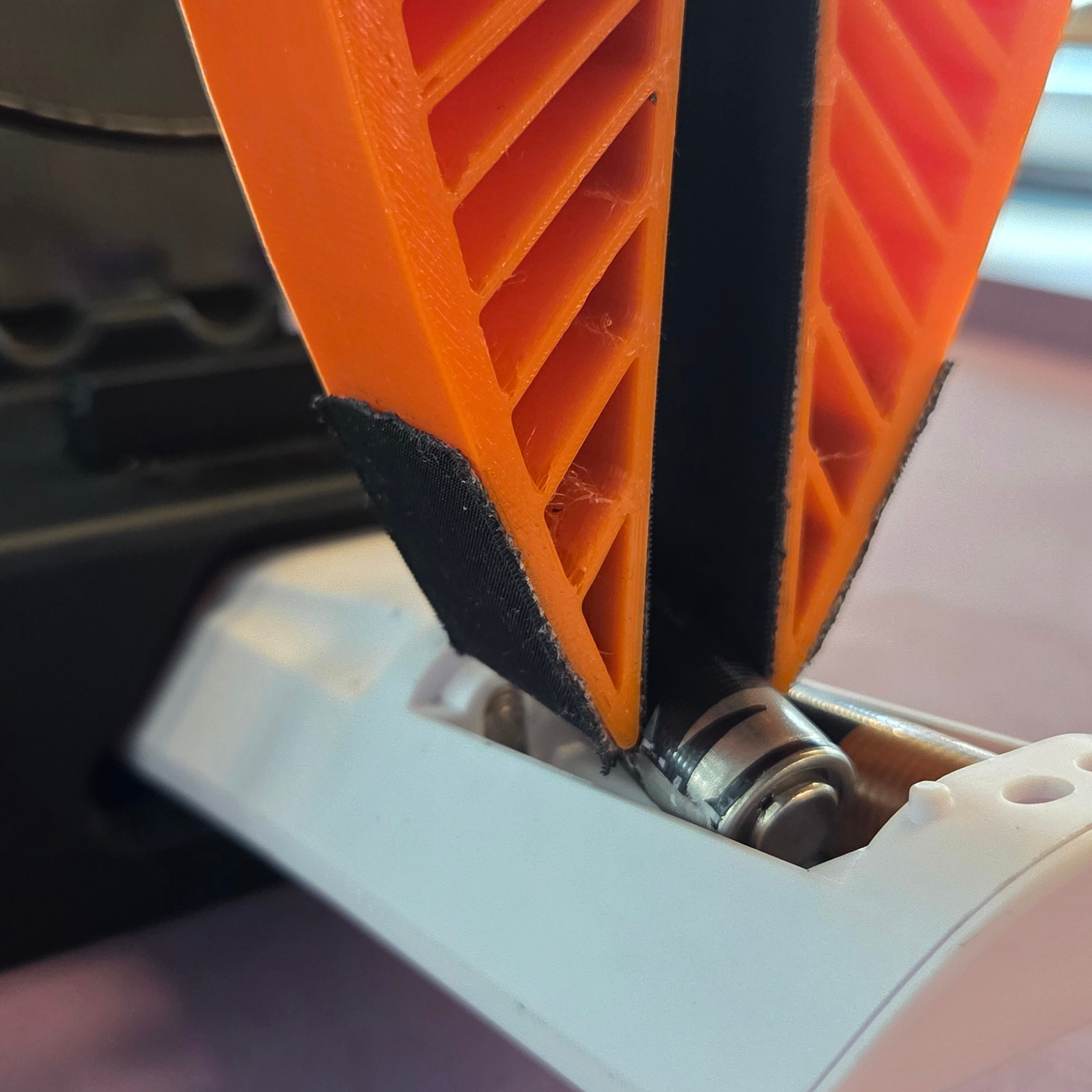}\end{subfigure} &
        \begin{subfigure}{0.155\linewidth}\centering\includegraphics[width=\linewidth]{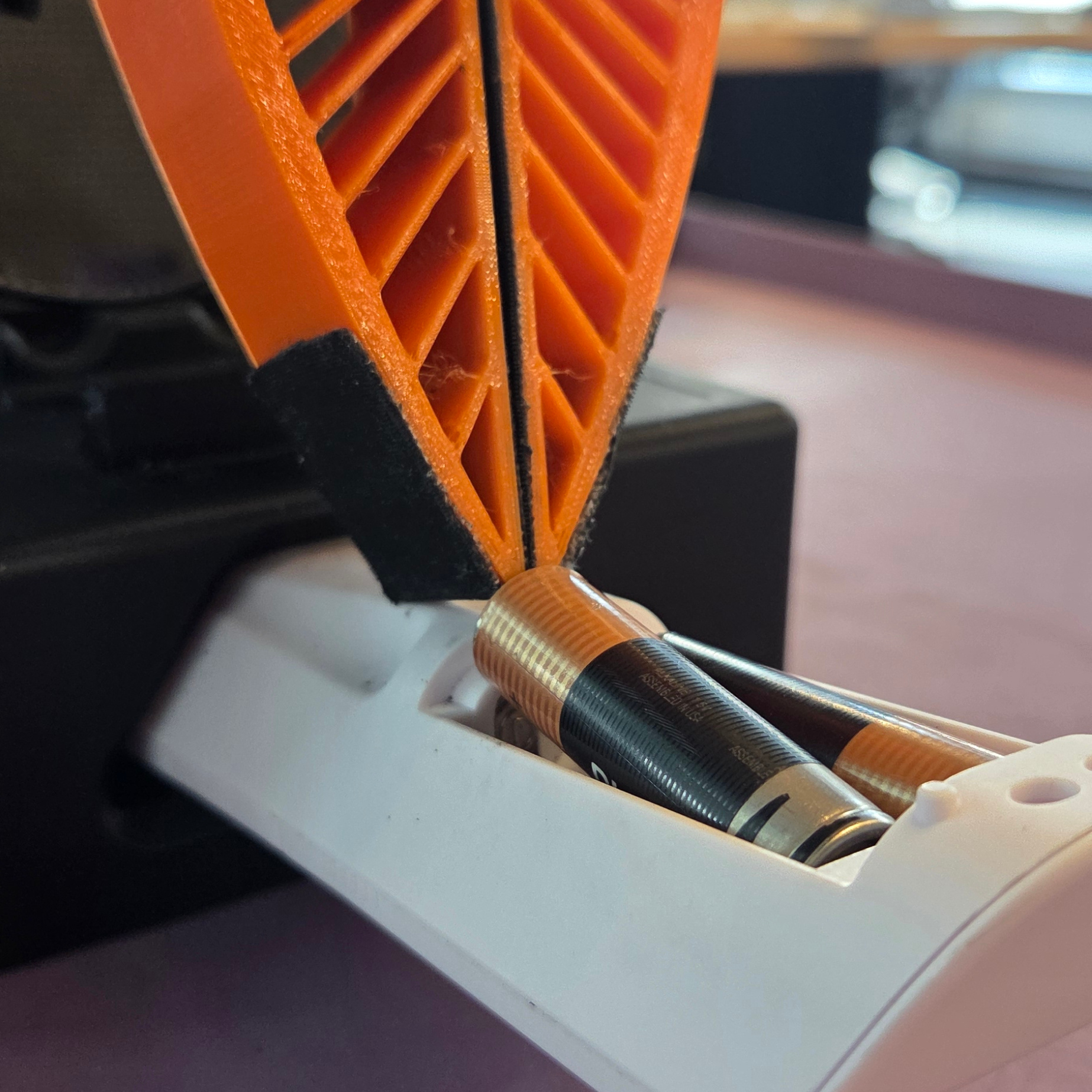}\end{subfigure} &
        \begin{subfigure}{0.155\linewidth}\centering\includegraphics[width=\linewidth]{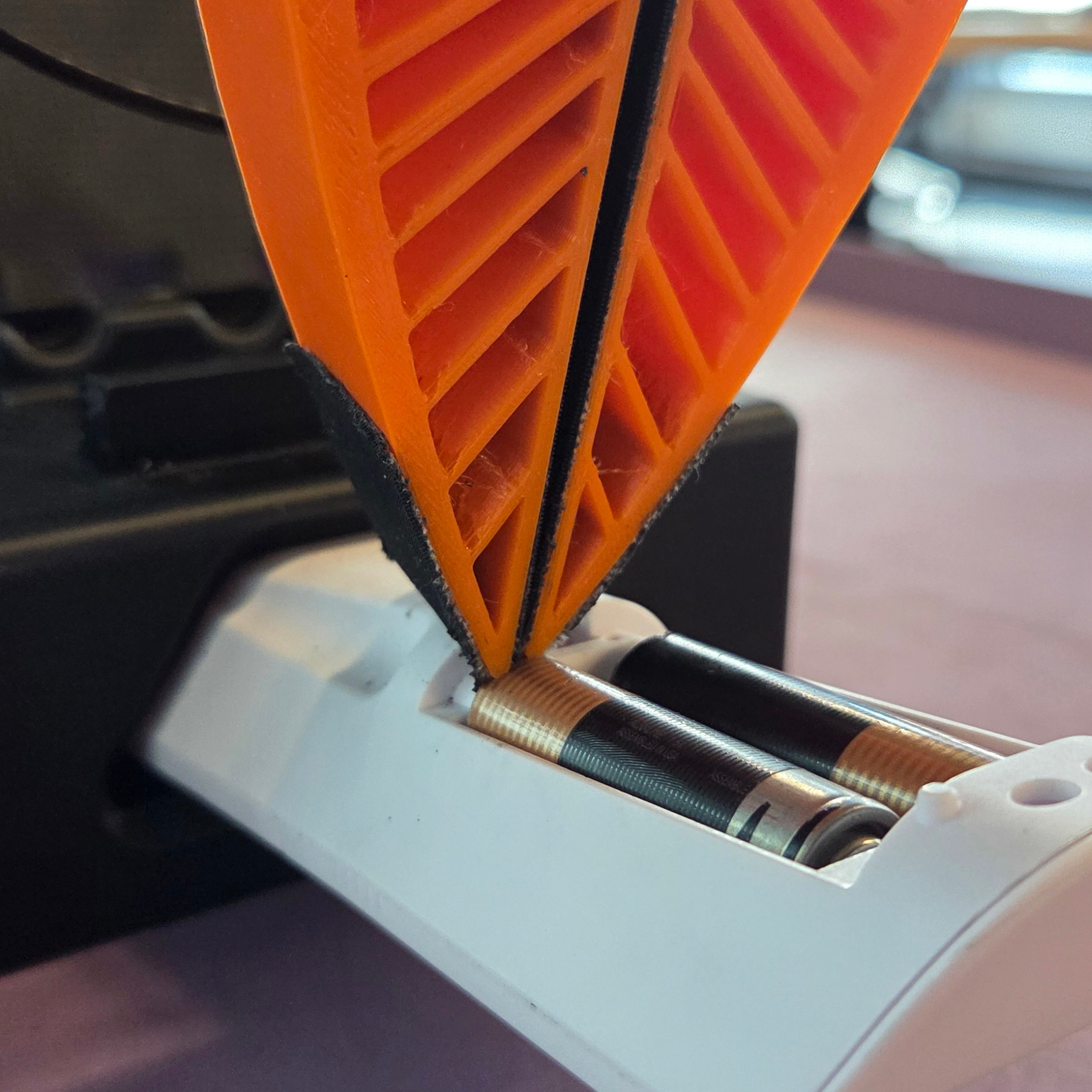}\end{subfigure} &
        \begin{subfigure}{0.155\linewidth}\centering\includegraphics[width=\linewidth]{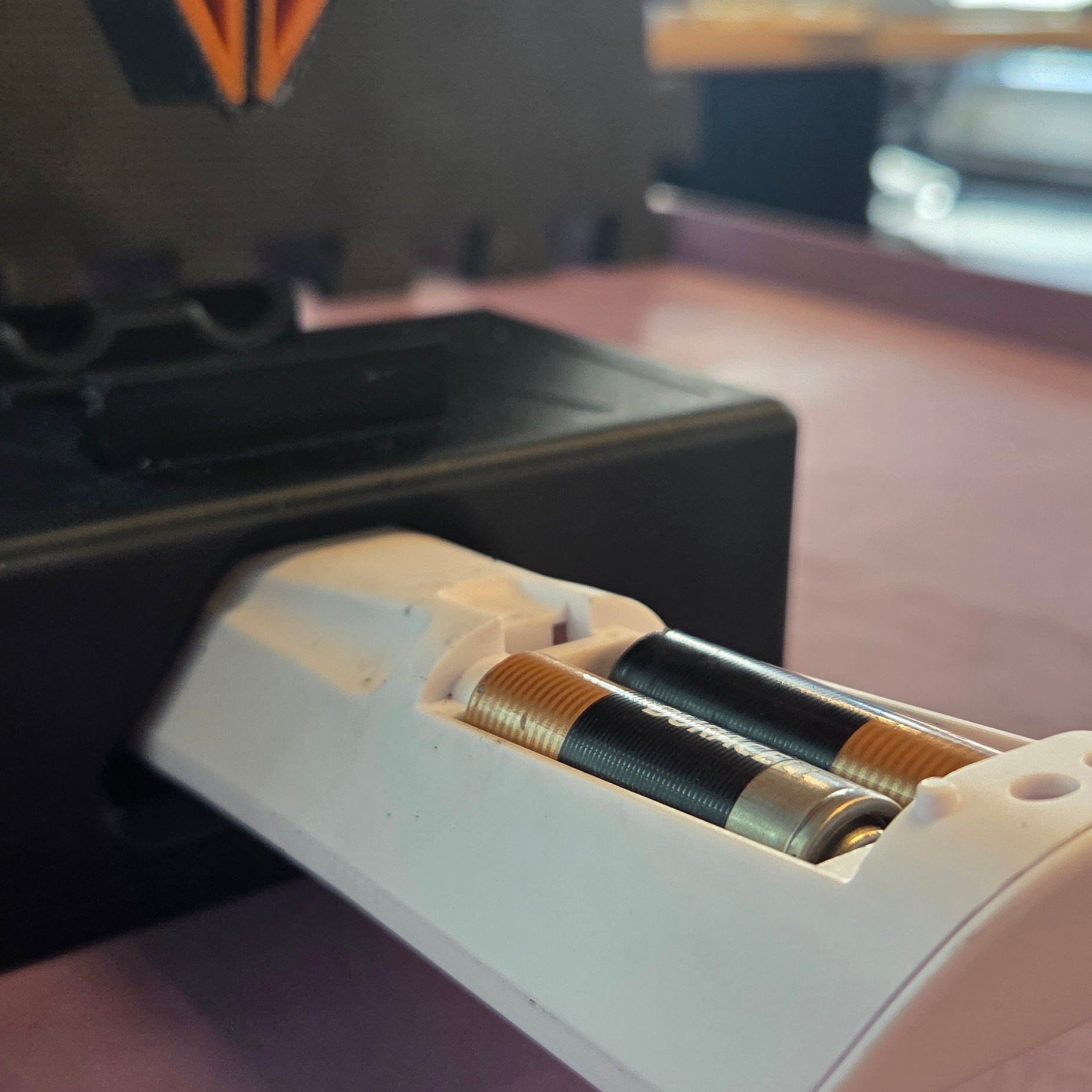}\end{subfigure}
    \end{tabular}
    \caption{\small\textbf{Policy Rollouts.} We evaluate three precise, contact-rich tasks, including NIST pulley routing (\textit{top)}, pipe insertion (\textit{middle}), and spring-loaded battery insertion (\textit{bottom)}.
}
    \label{fig:task_rollouts}
    \vspace{-6mm}
\end{figure*}

\textbf{Tasks.} We evaluate our method on three tasks (Figure~\ref{fig:task_rollouts}) that require precise geometric alignment and stable contact handling under external forces. These tasks highlight the need for distinguishing \textit{observed} and \textit{desired} actions:

\begin{itemize}[nosep]
\item \textit{NIST Pulley Routing.} The robot must grasp a deformable O-ring and route it around small and large pulleys. Adapted from NIST Assembly Task Board \#2~\cite{nist_assembly_task_boards}, this requires maintaining constant O-ring tension during precise end-effector movements.
\item \textit{Pipe Insertion.} The robot must grasp a pipe and insert it into a tight-tolerance opening. This requires reliable grasping, 6-DoF alignment, and force-aware insertion.
\item \textit{Battery Insertion.} The robot picks up a AA battery and inserts it into a spring-loaded compartment. This tests end-effector control under external forces, as the spring must be kept compressed while the battery is seated.
\end{itemize}

\textbf{Baselines.}
We evaluated the policies on a Franka FR3 7-DoF arm running a Cartesian impedance controller. All evaluated methods are based on the architecture described in Section~\ref{sec:model_arch}, differing only in their supervision strategy and training schedule. Similar to Chi et al. \cite{chi2024universal}, we used the relative end-effector and absolute gripper width formulation, and an action horizon of 24 steps. 

\begin{itemize}[nosep]
\item \textit{{\tt Base} Policy (Handheld).} We train exclusively on handheld DM-UMI demonstrations (observed actions). 

\item \textit{{\tt Base} Policy (Teleoperated).} We train exclusively on teleoperated demonstrations (desired actions) as an upper bound.

\item \textit{Naive Mix.} We train the {\tt Base} policy on a uniformly sampled combination of handheld and targeted teleoperated demonstrations.

\item \textit{\acronym \ (Ours).} We train the \acronym \ model on the {\tt base} and {\tt support} datasets following the procedure described in Section \ref{sec:model_arch}.
\end{itemize}

\textbf{NIST Pulley Routing Results.} 
This task evaluates spatial robustness during loaded routing, where the robot must maintain O-ring tension while moving around the pulley fixture. As shown in Table~\ref{tab:nist_pulley}, the {\tt Base} Policy (Handheld) succeeds in 44.0\% of trials, but its performance drops sharply in the top and bottom fixture positions. 
%These configurations exposed a platform-level limitation; high-stiffness tracking frequently triggered Franka controller faults, while smoothing the commanded motion reduced faults at the cost of degraded tracking. Similar safety faults can occur for lower stiffness teleoperation and {\tt support} data. - commented cause it got us in trouble without the table with faulted trials, but if you feel strongly add it back in.
The Naive Mix baseline fails because mixing observed and desired labels leads to imprecise manipulation; in practice, the policy often touches the top of the pulley and misses the routing groove. In contrast, \acronym~routes to the support expert for loaded routing and executes this segment with lower stiffness, improving success to 76.0\%,  approaching the full teleoperated policy at 84.0\%. These results suggest that targeted support data can recover much of the benefit of teleoperation while avoiding the need to teleoperate the full task.

\begin{table}[t]
    \centering
    \footnotesize
    \caption{\small\textbf{NIST Pulley Routing.} We report the results for 5 trials in each of the (C)enter, (L)eft, (R)ight, (T)op, and (B)ottom positions. We shift the rig $\pm 2$ cm in each direction.}
    \label{tab:nist_pulley}
    \renewcommand{\arraystretch}{0.9}
    \begin{tabular}{@{} l ccc @{}}
        \toprule
        & \multicolumn{3}{c}{Sub-task Performance (C-L-R-T-B)} \\
        \cmidrule(l){2-4}
        Method
        & \makecell{Full \\ (1.0)}
        & Score
        & \makecell{Task Success \\ Rate} \\
        \midrule
        \texttt{Base} Policy (Handheld)
        & 4--5--2--0--0
        & 11/25
        & 44.0\% \\
        Naive Mix
        & 0--0--0--0--0
        & 0/25
        & 0.0\% \\
        \acronym\ (Ours)
        & 4--4--4--4--3
        & 19/25
        & 76.0\% \\
        \texttt{Base} Policy (Teleoperated)
        & 5--5--4--4--3
        & 21/25
        & 84.0\% \\
        \bottomrule
    \end{tabular}
\end{table}

\textbf{Pipe Insertion Results.} This task tests geometric precision as small pose errors easily cause jamming. As shown in Table \ref{tab:pipe_insertion_detailed}, the {\tt Base} Policy (Handheld) frequently aligns the pipe but fails at insertion due to tracking errors near contact (13.3\% success). Naive Mix performs far below the {\tt base} handheld policy. This suggests that even with high stiffness, combining observed and desired action spaces into a single policy produces inconsistent alignment and contact behavior. High stiffness can partially compensate for the mismatch in some cases, but it does not replace the need for separating the observed and desired action spaces. In contrast, \acronym~uses targeted switching to command desired actions during the final alignment and insertion phase. This quadruples the success rate to 50.0\%, substantially closing the gap toward the teleoperated upper bound (63.3\%).

\begin{table}[t]
    \centering
    \caption{
    \small\textbf{Pipe Insertion Evaluation.} We report the results for 10 trials in each of the (L)eft, (C)enter, and (R)ight positions. We shift the rig $\pm 2$ cm in each direction.
    }
    \label{tab:pipe_insertion_detailed}
    \footnotesize
    % Reduces the vertical row padding to 70% of the default
    \renewcommand{\arraystretch}{0.7} 
    \resizebox{\columnwidth}{!}{%
    \begin{tabular}{@{} l cccc cc @{}}
        \toprule
        & \multicolumn{4}{c}{Sub-task Performance (L--C--R)} & & \\
        \cmidrule(lr){2-5}
        \makecell{Method \\ (Points)}
        & \makecell{Full \\ (1.0)}
        & \makecell{Partial \\ Insert \\ (0.75)}
        & \makecell{$\geq$50\% Aligned \\ Not Inserted \\ (0.5)}
        & \makecell{Failed \\ Alignment \\ (0.0)}
        & Score
        & \makecell{Task Success \\ Rate} \\
        \midrule

        \texttt{Base} Policy (Handheld) 
        & 0--2--2 
        & 0--1--0 
        & 10--7--8 
        & 0--0--0 
        & 17.25/30 
        & 13.3\% \\

        Naive Mix 
        & 0--0--2 
        & 0--0--4 
        & 0--4--2 
        & 10--6--2 
        & 8/30 
        & 6.7\% \\
        
        \acronym\ (Ours) 
        & 3--7--5 
        & 1--2--1 
        & 4--0--4 
        & 2--1--0 
        & 22.00/30 
        & 50.0\% \\

        \texttt{Base} Policy (Teleoperated)
        & 3--7--9 
        & 1--1--1 
        & 5--0--0 
        & 1--2--0 
        & 23.75/30 
        & 63.3\% \\
        
        \bottomrule
    \end{tabular}%
    }
\end{table}

\begin{table}[t]
    \centering
    \caption{\small\textbf{Spring-Loaded Battery Evaluation.} We report the results for 10 trials in each of the (L)eft, (C)enter, and (R)ight positions. We shift the rig $\pm 1$ cm in each direction.
    }
    \label{tab:spring_loaded_battery_insertion}
    \footnotesize
    % Reduces the vertical row padding to 70% of the default to shrink height by 30%
    \renewcommand{\arraystretch}{0.7}
    % Resizes the table to exactly the width of the text column
    \resizebox{\columnwidth}{!}{%
    \begin{tabular}{@{} l ccccc cc @{}}
        \toprule
        & \multicolumn{5}{c}{Sub-task Performance (L--C--R)} & & \\
        \cmidrule(lr){2-6}
        \makecell{Method \\ (Points)}
        & \makecell{Full \\ (1.0)}
        & \makecell{Not Fully \\ Inserted \\ (0.75)}
        & \makecell{Improperly \\ Placed \\ (0.5)}
        & \makecell{Ejected or \\ Dropped \\ (0.25)}
        & \makecell{Missed \\ Pick \\ (0.0)}
        & Score
        & \makecell{Task Success \\ Rate} \\
        \midrule

        \texttt{Base} Policy (Handheld)
        & 0--3--0 & 1--0--1 & 7--1--2 & 1--2--4 & 1--4--3 & 11.25/30 & 10.0\% \\
        Naive Mix 
        & 0--0--0 & 0--0--0 & 0--0--0 & 3--4--4 & 7--6--6 & 2.75/30 & 0.0\% \\
        \acronym\ (Ours)
        & 3--4--3 & 4--4--1 & 1--1--2 & 1--1--2 & 1--0--2 & 19.75/30 & 33.3\% \\
        \texttt{Base} Policy (Teleoperated)
        & 2--7--3 & 3--1--5 & 1--0--0 & 2--0--0 & 2--2--2 & 19.75/30 & 40.0\% \\

        \bottomrule
    \end{tabular}%
    \vspace{-6mm}
    }
\end{table}

\textbf{Spring-Loaded Battery Insertion Results.} This task demonstrates that action mismatch is primarily localized to contact phases (Table \ref{tab:spring_loaded_battery_insertion}). Although the Naive Mix picks up the battery roughly 40\% of the time, it exhibits a systematic lateral bias that shifts the battery rightward during transport, resulting in an insertion failure. This suggests that directly mixing observed and desired labels can distort the learned action distribution even before the final contact phase. The {\tt Base} Policy (Handheld) manages free-space transport well but struggles against the spring terminal, yielding a 10.0\% success rate. By preserving this free-space competence while using {\tt support} demonstrations to correct the delicate insertion phase, \acronym~more than triples the success rate to 33.3\%. Notably, \acronym~boosts the partial credit score from 11.25 to 19.75 and more unsuccessful trials get closer to completion (e.g., reaching the ``Not Fully Inserted'' stage). %The remaining gap to full teleoperated success (40.0\%) confirms that the remaining challenge is in regulating final contact forces.

\textbf{Statistical Analysis.} We compute Wilson 95\% confidence intervals over rollout success and two-sided Fisher exact tests comparing BRIDGE against the handheld and full-teleoperation baselines. BRIDGE improves over handheld supervision on pulley routing (76.0\% vs. 44.0\%, $p=0.042$) and pipe insertion (50.0\% vs. 13.3\%, $p=0.005$); the battery improvement (33.3\% vs. 10.0\%) does not reach $p<0.05$ at the evaluated trial count with $p=0.057$. BRIDGE is not statistically separable from full teleoperation on pulley, pipe, or battery ($p=0.73, 0.44, 0.79$, respectively). These intervals quantify rollout uncertainty rather than variation across independently trained policy seeds.

\begin{figure}[t]
    \centering
    \includegraphics[width=\linewidth]{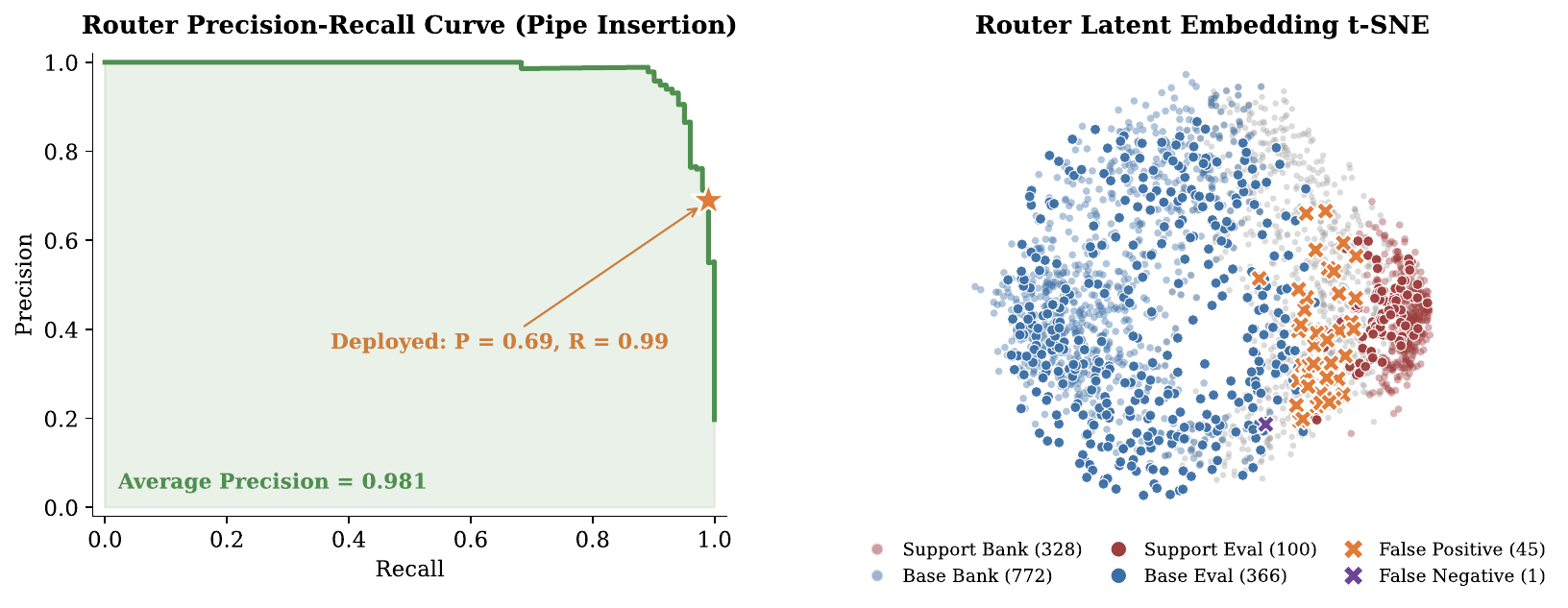}
    \caption{\small\textbf{Router Analysis.} We visualize the precision-recall curve for the pipe insertion task (left). The deployed router achieves 99.0\% recall and 69.0\% precision, favoring early {\tt support} activation over missed handoffs. Computed t-SNE latent embeddings demonstrate clear separation between {\tt base} and {\tt support} states, yielding only a single false negative (right).
}
    \label{fig:gate_analysis}
\end{figure}

\textbf{Router Analysis.} 
We evaluate our router on the challenging pipe insertion task. We use a held-out teleoperated dataset with manually labeled {\tt support} phases to prevent the router evaluation from artificially inflating performance by exploiting dataset or embodiment artifacts. On this dataset, our MLP router achieves 99.0\% recall and 69.0\% precision. High recall is desirable to prevent the {\tt base} policy from incorrectly handling insertion alignment. However, because the transition between {\tt base} and {\tt support} phases is inherently ambiguous, frame-level labels are imprecise near the boundary. Figure \ref{fig:gate_analysis} shows that false positives concentrate near the {\tt support} manifold boundary (representing transition states) rather than unrelated {\tt base} states. This aligns with our nearest-neighbor design, where the {\tt support} bank covers the {\tt support}-state distribution, naturally restricting successful routing to this manifold rather than requiring extrapolation to disjoint regions. 
%We ablate hyperparameters used to train the router in the supplement.

\begin{figure}[t]
    \centering
    \begin{tikzpicture}
        \begin{axis}[
            width=0.9\columnwidth,
            height=0.62\columnwidth,
            xmin=-10,
            xmax=240,
            ymin=0,
            ymax=80,
            xlabel={Additional Demonstrations Beyond 100 Handheld},
            ylabel={Task Success Rate (\%)},
            grid=major,
            tick label style={font=\footnotesize},
            label style={font=\small},
            legend style={
                at={(0.5,1.04)},
                anchor=south,
                legend columns=2,
                font=\footnotesize,
                draw=none,
                fill=none
            },
            mark size=2.8pt,
        ]

        % Handheld scaling beyond the first 100 handheld demos
        \addplot+[
            thick,
            mark=square*,
        ] coordinates {
            (0,13.3)
            (100,23.3)
            (228,26.7)
        };
        \addlegendentry{Additional \texttt{Base}}

        % Targeted partial teleop added to the same 100 handheld base
        \addplot+[
            thick,
            mark=triangle*,
        ] coordinates {
            (0,13.3)
            (81,50.0)
            (200,73.3)
        };
        \addlegendentry{Additional \texttt{Support}}

        \end{axis}
    \end{tikzpicture}
    \caption{
    \textbf{Impact of Scaling Targeted \texttt{Support} Data.} Starting from the same 100-demo \texttt{base} policy, incorporating targeted partial teleoperation demonstrations improves pipe insertion success substantially more than simply adding more handheld demonstrations. While adding over 200 additional handheld demonstrations yields only marginal improvements, utilizing our targeted partial teleoperation approach increases task success to over 70\%. 
    }
    \label{fig:scaling_graph}
    \vspace{-6mm}
\end{figure}
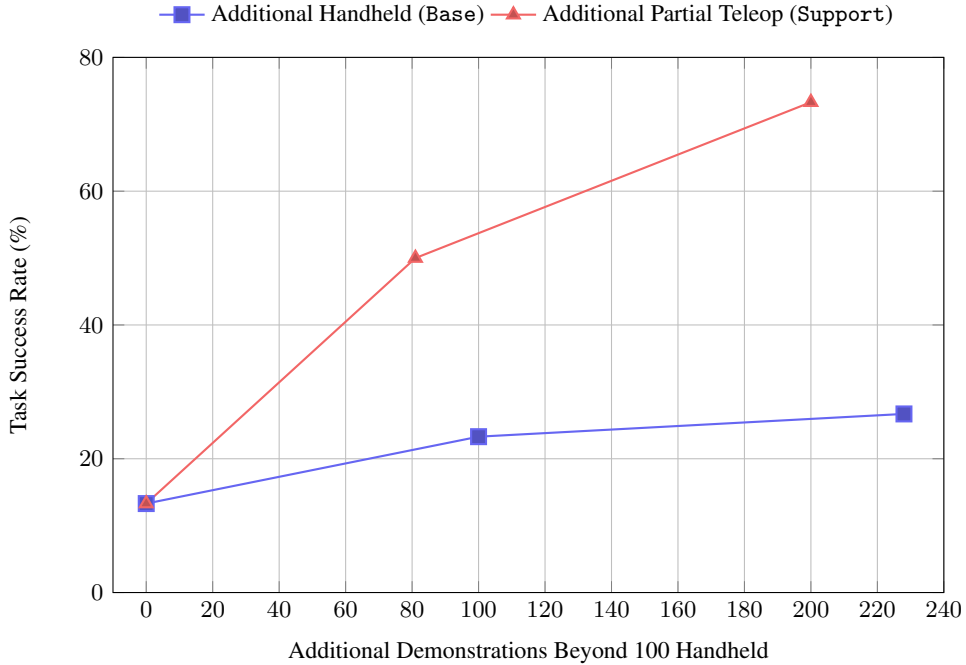

\begin{table}[t]
    \centering
    \caption{
\small\textbf{Pipe Insertion Scaling Ablations.} We report the results for 10 trials in each of the (L)eft, (C)enter, and (R)ight positions. We shift the rig $\pm 2$ cm in each direction.
    }
    \label{tab:pipe_insertion_ablation}
    \footnotesize
    % Reduces the vertical row padding to 70% of the default
    \renewcommand{\arraystretch}{0.7} 
    \resizebox{\columnwidth}{!}{%
    \begin{tabular}{@{} l cccc cc @{}}
        \toprule
        & \multicolumn{4}{c}{Sub-task Performance (L--C--R)} & & \\
        \cmidrule(lr){2-5}
        \makecell{Method \\ (Points)}
        & \makecell{Full \\ (1.0)}
        & \makecell{Partial \\ Insert \\ (0.75)}
        & \makecell{$\geq$50\% Aligned \\ Not Inserted \\ (0.5)}
        & \makecell{Failed \\ Alignment \\ (0.0)}
        & Score
        & \makecell{Task Success \\ Rate} \\
        \midrule

        \texttt{Base} Policy (Full Teleop, 60 demos)
        & 3--7--9 
        & 1--1--1 
        & 5--0--0 
        & 1--2--0 
        & 23.75/30 
        & 63.3\% \\

        \texttt{Base} Policy (Full Teleop, 120 demos)
        & 9--10--9 
        & 0--0--0 
        & 1--0--1 
        & 0--0--0 
        & 29.00/30 
        & 93.3\% \\

        \midrule 
        
        \texttt{Base} Policy (Handheld, 100 demos)
        & 0--2--2 
        & 0--1--0 
        & 10--7--8 
        & 0--0--0 
        & 17.25/30 
        & 13.3\% \\
        
        \texttt{Base} Policy (Handheld, 200 demos)
        & 0--4--3 
        & 0--0--0 
        & 10--6--7 
        & 0--0--0 
        & 18.50/30 
        & 23.3\% \\

        \texttt{Base} Policy (Handheld, 328 demos)
        & 0--5--3 
        & 0--0--0 
        & 10--5--7 
        & 0--0--0 
        & 19.00/30 
        & 26.7\% \\

        \midrule 
        
        \acronym~(100 Handheld + 81 Partial Teleop)
        & 3--7--5 
        & 1--2--1 
        & 4--0--4 
        & 2--1--0 
        & 22.00/30 
        & 50.0\% \\
        
        \acronym~(100 Handheld + 200 Partial Teleop)
        & 6--8--8 
        & 2--2--0 
        & 2--0--2 
        & 0--0--0 
        & 27.00/30 
        & 73.3\% \\

        \bottomrule
    \end{tabular}%
    }
    \vspace{-6mm}
\end{table}

\textbf{Preliminary Exploration of Data Scaling.} We investigate the impact of handheld and targeted teleoperated demonstrations on the pipe insertion task. Table~\ref{tab:pipe_insertion_ablation} summarizes these initial scaling results. Training a base expert on 120 fully teleoperated demonstrations yields a 93.3\% success rate, significantly improving over the 63.3\% achieved by the 60 demonstration baseline. This high performance is expected; teleoperated data is collected directly on the evaluation robot and pipe insertion setup, effectively minimizing the embodiment gap.

Increasing the number of handheld demonstrations alone provides marginal gains. Doubling the original 100 demonstration dataset to 200 demonstrations achieves a 23.3\% full-insertion success rate and a score of 18.50/30. Scaling further to 328 UMI demonstrations slightly increases success to 26.7\% and the score to 19.00/30. These results suggest that the primary failure mode is not insufficient coverage of nominal UMI behavior, but rather imperfect trajectory tracking under contact-rich regimes. In contrast, scaling the partial teleoperation support data from 81 to 200 demonstrations yields substantial improvements. By combining the original 100 demonstration handheld {\tt base} expert with a {\tt support} expert trained on this expanded dataset, \acronym~achieves a 73.3\% full-insertion success rate and a score of 27.00/30. This is a marked improvement over the 50.0\% success and 22.00/30 score achieved with the original 81 demonstration {\tt support} set. This improvement stems from converting ``aligned-but-not-inserted'' states into full insertions.

\textbf{Cross-Robot Teleoperation Transfer.}
The 120 demonstration full teleoperation pipe insertion policy with a success rate of 93.3\% represents an in-domain upper bound. To examine hardware sensitivity, we evaluated it on a second Franka FR3 (Robot B) for 10 trials with the pipe at the Center position. Full task success rate dropped sharply to 10\%, with most trials reaching only partial alignment. This outcome aligns with the known sensitivity of policies to variations in robot calibration, controller dynamics, and minor setup biases. We then augmented the handheld {\tt base} policy with 50 targeted support demonstrations collected on Robot B in about 15 minutes wall time. \acronym~achieved a full task success rate of 30\% with a task completion score of $7.00/10$ (3 full, 4 partial, 2 aligned, 1 failure). While further investigation is required, this preliminary result suggests that a reusable handheld task scaffold can help recover cross-robot performance through targeted teleoperation, reducing reliance on full task teleoperation.

\begin{table}[t]
\centering
\footnotesize
\setlength{\tabcolsep}{2.2pt}
\caption{\textbf{Ablation on Compliance and Support Supervision.} We ablate the impedance schedule and support-head supervision on the pulley task below. Lowering the {\tt Base} Policy (Handheld) from HIGH to LOW stiffness reduces success from 44\% to 24\%, and scheduling stiffness with BRIDGE's router (State-Gated Stiffness) recovers only 28\%, compared to BRIDGE's 76\%. Training the support head on observed rather than desired actions (OBS Support) reaches 20\%. Importantly, this suggests that BRIDGE's gain comes from routing between policies rather than from compliance alone. We denote the {\tt Base} policy with B and the {\tt Support} policy with S, where HIGH uses 1600 N/m and 60 Nm/rad and LOW uses 800 N/m and 30 Nm/rad, both critically damped. All other controller parameters are identical across methods.}
\label{tab:fault_accounting}
\begin{tabular}{lcc}
\toprule
Method & Compliance & Task Success Rate \\
\midrule
% runs reported in paper
Teleoperation  & LOW & 21/25 (84\%) \\
BRIDGE   & B:HIGH, S:LOW & 19/25 (76\%) \\
Handheld & HIGH & 11/25 (44\%) \\
\midrule
Handheld  & LOW  & 6/25 (24\%) \\ % New run
Handheld, State-Gated Stiffness & HIGH/LOW & 7/25 (28\%) \\ % new run
BRIDGE, OBS Support & B:HIGH, S:LOW & 5/25 (20\%) \\ % new run
\bottomrule
\end{tabular}
\vspace{-3mm}
\end{table}

\textbf{Ablation on Compliance and Support Supervision.} We ablate BRIDGE's controller stiffness, and action supervision used to train the {\tt support} head on the pulley task (Table ~\ref{tab:fault_accounting}). Lowering the controller stiffness to LOW and using the {\tt Base} Policy (Handheld) drops success from 44.0\% to 24.0\%, and scheduling its impedance with BRIDGE's router (Handheld, State-Gated Stiffness) reaches only 28.0\%, compared to BRIDGE's 76.0\%. This suggests that the change in controller stiffness when switching to the {\tt support} head is not the sole reason for improvement. We find that the source of action supervision matters as well. We retrain the support head on observed rather than desired actions (BRIDGE, OBS Support), using the same teleoperated support trajectories with all else identical. This variant achieves 5/25 (20.0\%) compared to BRIDGE's 19/25 (76.0\%).

\textbf{Ablation on Router Labels.} Training the task router on raw source labels rather than kNN pseudo-labels achieves 100\% recall and 62.8\% precision ($\eta$=0.5) on the held-out test set. However, this router succeeds on only 9/25 (36\%) trials on the robot, compared to 19/25 (76\%) for our kNN pseudo-label router and 22/25 (88\%) for an oracle router that we construct by manually switching between policies, suggesting that $\epsilon$-filtering the kNN pseudo-labels to remove overlap is critical.

\begin{figure*}[t]
    \centering
    
    % Battery Trajectory
    \begin{subfigure}[t]{0.32\linewidth}
        \centering
        \includegraphics[
            width=\linewidth,
            trim={0cm 0cm 0cm 1cm},
            clip
        ]{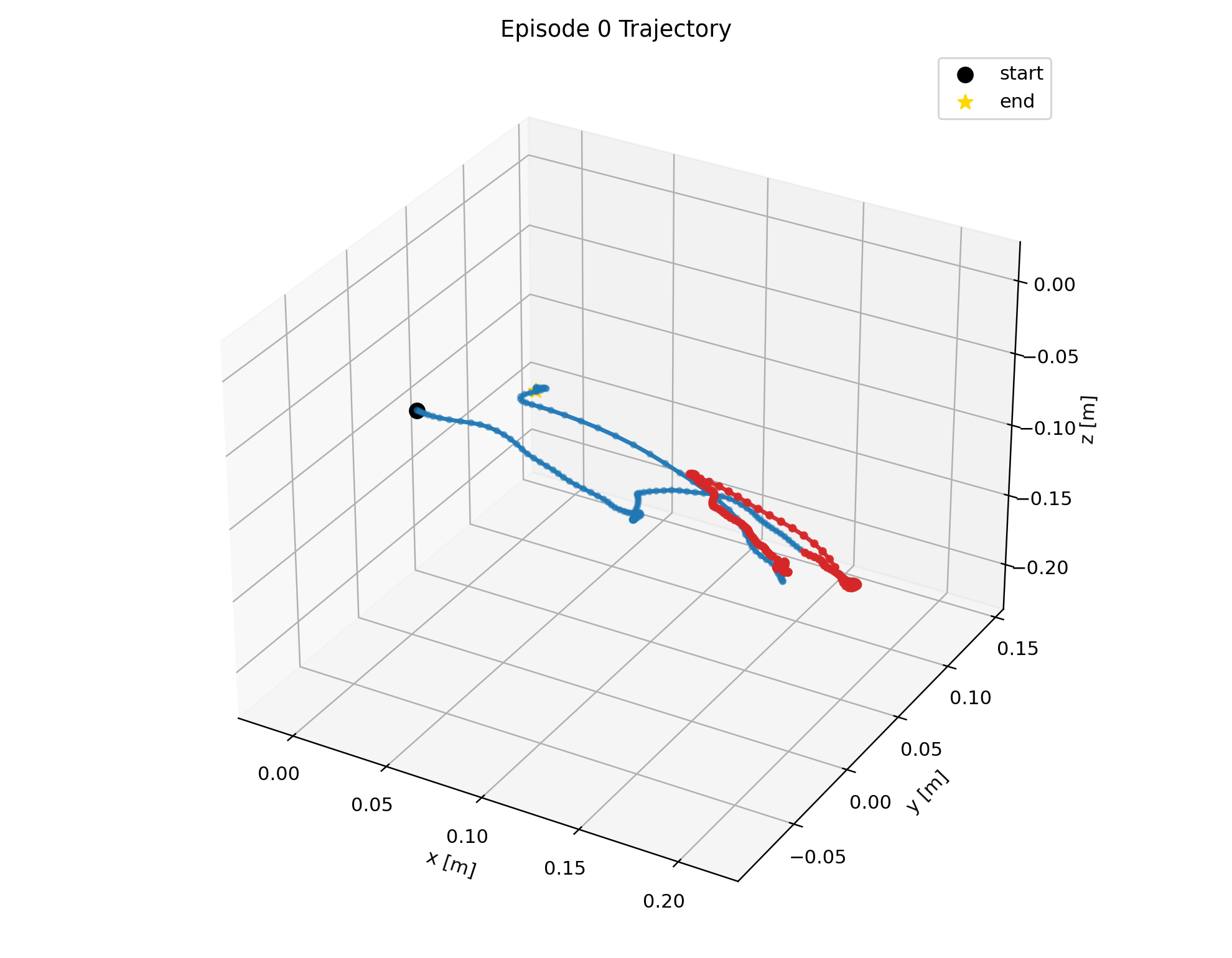}
        \caption{}
        \label{fig:traj_battery}
    \end{subfigure}\hfill
    % Pipe Trajectory
    \begin{subfigure}[t]{0.32\linewidth}
        \centering
        \includegraphics[
            width=\linewidth,
            trim={0cm 0cm 0cm 1cm},
            clip
        ]{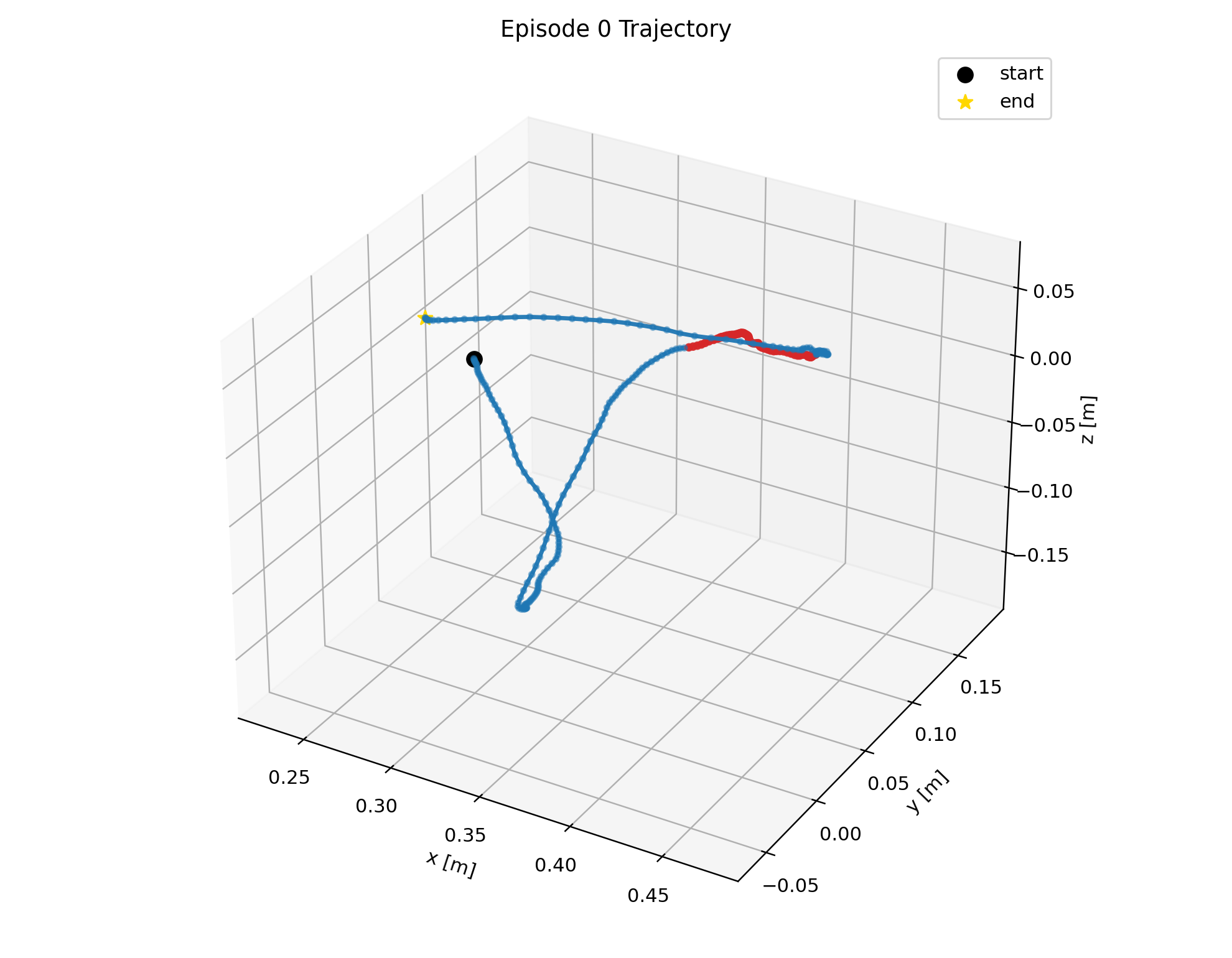}
        \caption{}
        \label{fig:traj_pipe}
    \end{subfigure}\hfill
    % Pulley Trajectory
    \begin{subfigure}[t]{0.32\linewidth}
        \centering
        \includegraphics[
            width=\linewidth,
            trim={0cm 0cm 0cm 1cm},
            clip
        ]{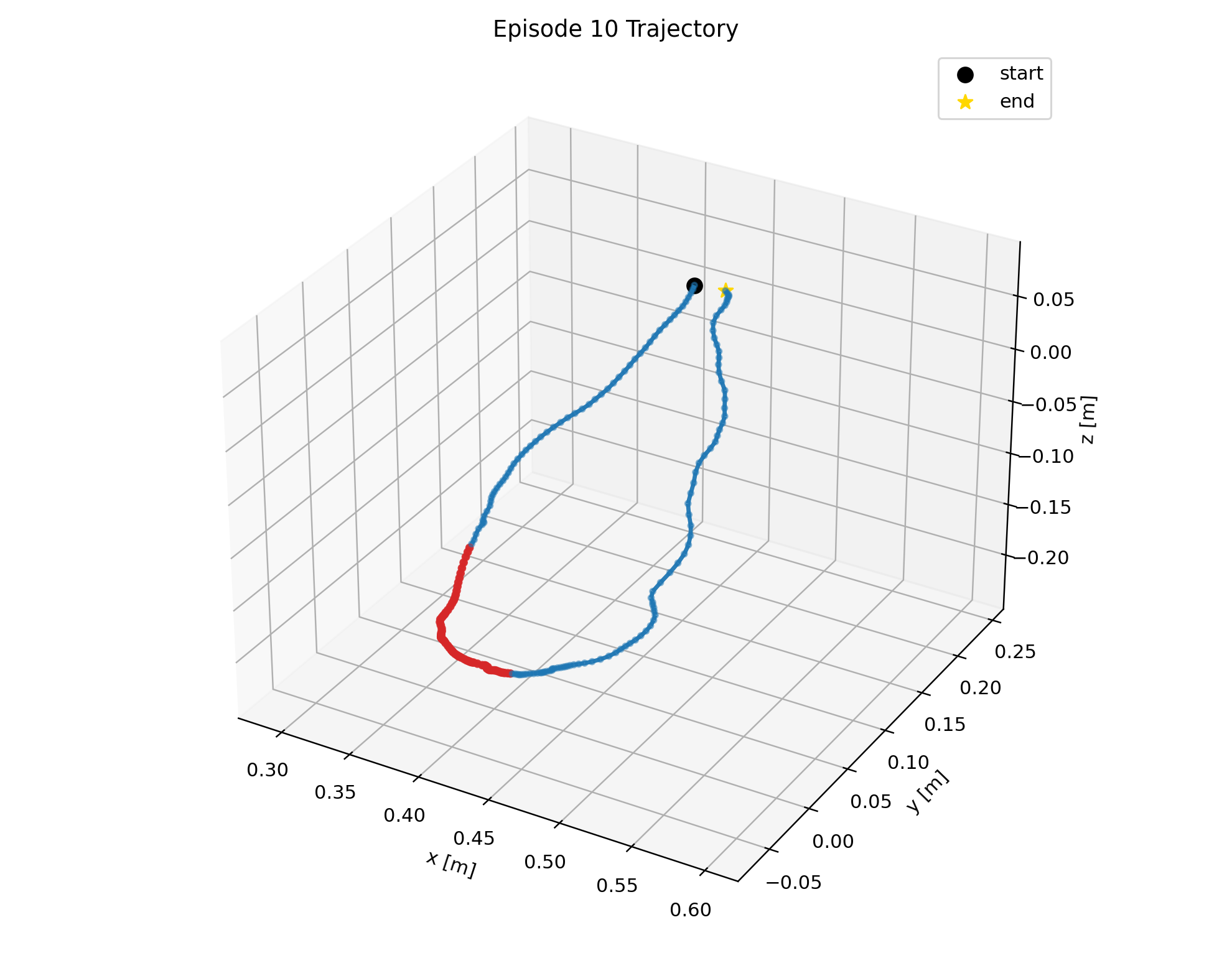}
        \caption{}
        \label{fig:traj_pulley}
    \end{subfigure}

    \caption{
        \small\textbf{Trajectory Support Coverage Across Tasks.} As illustrated by the red highlighted segments, the median trajectory {\tt support} coverage spans 14.9\% of the trajectory length for battery insertion (a), 3.4\% for pipe insertion (b), and 7.4\% for NIST pulley routing (c).
    }
    \label{fig:combined_traj_support}
    \vspace{-6mm}
\end{figure*}

\textbf{Visualization of Support Regions.}
Figure~\ref{fig:combined_traj_support} illustrates the 3D end-effector trajectories for battery insertion, pipe insertion, and NIST pulley routing. Each plot displays a single episode's full trajectory, originating at the start position (marked by a black dot) and concluding at the end position (marked by a yellow star). Within each blue trajectory path, the critical phase (identified as the {\tt support} segment) is highlighted in red. Specifically, the median {\tt support} segment accounts for 14.9\% of the total trajectory length in the battery insertion task, 3.4\% in the pipe insertion task, and 7.4\% in the NIST pulley routing task.

\textbf{Limitations and Future Work.} Our approach for identifying which task phases require {\tt support} data requires training and evaluating a {\tt base} policy. Future work should formalize this criterion so that it can be estimated automatically from the training data. Rather than inferring {\tt support} regions from a trained {\tt base} model, UMI-style devices instrumented with force-torque sensing \cite{choi2026wild} may allow us to identify them directly.

\section{Conclusion}
In this work, we address the embodiment gap between handheld data collection and teleoperation through the lens of \textit{action validity}. We demonstrate that handheld data, which provides observed actions, offers valid supervision for error-tolerant phases but can fail during contact-sensitive phases that require dynamically feasible desired actions. To improve data collection efficiency, we propose \method\ (\acronym), a mixture of diffusion policy experts that uses scalable handheld data to learn the general task scaffold and sparse teleoperated data to target contact-rich bottlenecks. Across our evaluated tasks, this sparse teleoperated data covers less than 15\% of the end-effector path length, yet \acronym~ achieves at least 79\% of the full teleoperation upper bound success rate. By separating observed and desired action heads, \acronym\ provides a practical mechanism for specializing behavior across task phases.

\bibliographystyle{IEEEtran}
\bibliography{egbib.bib}

\end{document}